\documentclass[sensors,article,accept,pdftex,moreauthors]{Definitions/mdpi} 

\usepackage{graphicx} 
\usepackage{float}    
\usepackage{caption}
\usepackage{soul}

\usepackage{amsmath}
\usepackage{makecell}

\firstpage{1} 
\pubvolume{25}
\issuenum{1}
\articlenumber{0}
\pubyear{2025}
\copyrightyear{2025}
\externaleditor{Giovanni Saggio} 
\datereceived{13 May 2025} 
\daterevised{13 June 2025} 
\dateaccepted{25 June 2025 } 
\datepublished{ } 
\hreflink{https://doi.org/} 
       
\Title{Reducing Label Dependency in Human Activity Recognition with Wearables: From Supervised Learning to Novel Weakly Self-Supervised Approaches}

\TitleCitation{Reducing Label Dependency in Human Activity Recognition with Wearables: From Supervised Learning to Novel Weakly Self-Supervised Approaches}

\Author{Taoran Sheng 
$^{1,*}$\orcidA{} and Manfred Huber$^{1}$\orcidB{}}

\AuthorNames{Taoran Sheng and Manfred Huber}

\isAPAStyle{%
       \AuthorCitation{Lastname, F., Lastname, F., \& Lastname, F.}
         }{%
        \isChicagoStyle{%
        \AuthorCitation{Lastname, Firstname, Firstname Lastname, and Firstname Lastname.}
        }{
        \AuthorCitation{Sheng, T.; Huber, M.}
        }
}

\address[1]{%
$^{1}$ \quad The University of Texas at Arlington; Arlington, TX 76019, USA; tsheng0122@gmail.com 
\\
$^{1}$ \quad The University of Texas at Arlington; Arlington, TX 76019, USA; huber@cse.uta.edu 
}

\corres{Correspondence: {tsheng0122@gmail.com}}

\abstract{Human activity recognition (HAR) using wearable sensors has advanced through various machine learning paradigms, each with inherent trade-offs between performance and labeling requirements. While fully supervised techniques achieve high accuracy, they demand extensive labeled datasets that are costly to obtain. Conversely, unsupervised methods eliminate labeling needs but often deliver suboptimal performance. This paper presents a comprehensive investigation across the supervision spectrum for wearable-based HAR, with particular focus on novel approaches that minimize labeling requirements while maintaining competitive accuracy. We develop and empirically compare: (1) traditional fully supervised learning, (2) basic unsupervised learning, (3) a weakly supervised learning approach with constraints, (4) a multi-task learning approach with knowledge sharing, (5) a self-supervised approach based on domain expertise, and (6) a novel weakly self-supervised learning framework that leverages domain knowledge and minimal labeled data. Experiments across benchmark datasets demonstrate that: (i) our weakly supervised methods achieve performance comparable to fully supervised approaches while significantly reducing supervision requirements; (ii) the proposed multi-task framework enhances performance through knowledge sharing between related tasks; (iii) our weakly self-supervised approach demonstrates remarkable efficiency with just 10\% of labeled data. These results not only highlight the complementary strengths of different learning paradigms, offering insights into tailoring HAR solutions based on the availability of labeled data, but also establish that our novel weakly self-supervised framework offers a promising solution for practical HAR applications where labeled data are limited.}


\usepackage{mycomments3}
\begin{document}

\section{Introduction 
}

Wearable-sensor-based HAR is vital for applications in health monitoring, human--computer interaction, and many other areas. However, most current HAR techniques heavily rely on labeled training data, which are often scarce and costly to annotate. This reliance on labeled data presents significant challenges: fully supervised deep learning methods require large labeled datasets, while purely unsupervised techniques typically underperform due to the lack of informative cues. As a result, this labeling challenge hinders the real-world deployment of HAR systems.

To address this challenge, we explored various deep representation learning approaches to reduce label reliance in HAR with wearables. Specifically, we reviewed different learning paradigms within representation learning. We then carefully designed and tailored these approaches by introducing novel model architectures, incorporating domain knowledge, and adopting diverse training methods to effectively address label dependency. Our study focused on evaluating how these modifications impact the label dependency of these approaches. In particular, we explored the following approaches to reduce label reliance:

\begin{enumerate}[label=\roman*.]
 \item We 
 first developed a weakly supervised approach inspired by previous research in computer vision \cite{Koch2015SiameseNN,faceverification}. These prior works demonstrated that selective representation can be achieved using weakly supervised approaches that reduce the need for extensive labeled training data. Building on these findings, we incorporated the weakly supervised approach as a means to decrease the model's reliance on labeled data.

\item We next enhanced our weakly supervised approach by integrating multi-task learning to further reduce label reliance by leveraging shared knowledge between related tasks. While achieving favorable outcomes is feasible through learning solely on a single task, potentially superior results might be attainable through the inclusion of related tasks. This is because insights drawn from the training signals of correlated tasks can offer valuable information to improve the model's competence in the primary task. By sharing representations among correlated tasks, we can empower the model to achieve better overall performance. From a multi-task standpoint \cite{MTL, 10.1145/3328932, Sheng2020Multitask}, HAR and person identification exhibit a strong connection. Integrating them within a single weakly supervised multi-task model is a logical step and holds potential advantages for both tasks.

\item We also implemented a self-supervised learning approach that incorporates domain expertise to create supervisory signals from the intrinsic structure of the data. This method leverages the inherent patterns and semantics within sensor data, thereby circumventing the need for manual labeling. In the context of wearable sensors, self-supervised methods can utilize specialized knowledge about human activity movement patterns, biomechanics, and contextual information to construct pretext tasks. These tasks encourage the model to learn representations that capture meaningful aspects of the data without requiring explicit annotations. By integrating this domain understanding into pretext tasks, our approach effectively captures high-level abstractions from raw sensor data, leading to more robust and interpretable representations. Through this method, we address the label dependence challenge in HAR by deriving useful supervisory signals directly from the sensor data.

\item Finally, we developed a novel hybrid framework that combines weak supervision with self-supervision to further reduce label dependency. This weakly self-supervised approach integrates the complementary strengths of both paradigms, enabling the model to benefit from the structured guidance of weak supervision while simultaneously leveraging self-supervised signals from unlabeled data. To demonstrate the practical benefits of this innovative hybrid approach, we conducted experiments with extremely limited label availability (1\%, 5\%, and 10\% of labeled data), showing how our combined framework progressively improves performance as more labels become available.

\end{enumerate}

By evaluating and integrating various learning paradigms, we aim to advance HAR systems that are less reliant on extensive labeled data. Our comprehensive experimental evaluation compares our proposed approaches against both traditional supervised and unsupervised baselines as well as established approaches from prior research, providing insights into the trade-offs between supervision requirements and recognition performance. The remainder of this paper outlines the theoretical foundations and methodologies that guide our investigation, setting the stage for the comparative analysis of these strategies.


\section{Background} \label{bckgrnd}
In this section, we provide a comprehensive overview of the foundational concepts and techniques central to our study. We begin with the fundamentals of HAR, which involves identifying and categorizing activities based on data from wearable sensors. Next, we explore deep representation learning, a key aspect of modern machine learning that focuses on automatically extracting meaningful features from raw data. Finally, we examine the technical approaches employed in our research, including various deep learning architectures and specialized loss functions designed to reduce label reliance in HAR. These components form the theoretical and practical foundation upon which our proposed methods are built.

\subsection{Human Activity Recognition} \label{har}
Wearable sensor-based HAR approaches generally fall into two main categories: methods using carefully designed handcrafted features, and those leveraging deep neural networks (DNNs) to automatically derive discriminative representations from raw sensor data.

\subsubsection{Handcrafted Features}
Handcrafted features are deliberately designed based on domain expertise, offering a manual approach to feature engineering. In HAR systems, these features are specifically tailored to capture relevant information from sensor data for accurate activity recognition. Various statistical features, such as mean, variance, and entropy, have been successfully integrated into models \cite{sbhar,transitionAware,Sheng2022consistency}. Additionally, researchers have employed wavelet transform-based features \cite{waveletFeature}, while He and Jin \cite{dctFeature} proposed features derived from the discrete cosine transform. These signal-based features have proven effective in HAR systems, demonstrating their capability to accurately capture crucial information for activity recognition.

\subsubsection{Learned Features}
In recent years, DNNs \cite{DLHAR, 10.1145/3550299} have significantly transformed HAR. DNNs enable automatic extraction of features from raw sensor data, eliminating the need for manual feature crafting. Morales and Roggen \cite{cnnLSTMHAR} designed a model that integrates Convolutional Neural Networks (CNNs) with Long Short-Term Memory (LSTM) components. While CNNs capture local temporal relationships, LSTM's memory states facilitate the understanding of broader time-scale dependencies. Abu Alsheikh et al. \cite{DBNHAR} introduced a hybrid approach combining a deep belief network with a Hidden Markov Model, effectively leveraging the strengths of both paradigms for feature extraction. More recently, Chen et al. \cite{chen2021modeling} proposed a framework where a deep network architecture utilizes stage distillation, progressively extracting more informative features from raw data, enhancing HAR performance. These DNN-based strategies excel in automated feature extraction but still require explicit labels for model training.

\subsubsection{Hybrid Features}
Beyond the traditional dichotomy between handcrafted and learned features, recent studies have pursued hybrid strategies that integrate the best of both worlds. \mbox{Qin et al. \cite{QIN202080},} for instance, incorporated engineered wavelet features organized as an image input. These features were then processed through a convolutional network to autonomously extract and abstract the engineered attributes, resulting in enhanced representation. This hybrid approach combines the interpretability of handcrafted features with the automation power of DNNs, offering increased flexibility to the learning system. However, similar to other approaches, this hybrid strategy still relies on explicit labels for effective training.

For a comprehensive understanding of various aspects and methodologies in HAR, we refer readers to survey papers by Minh Dang et al. \cite{MINHDANG2020107561}, Wang et al. \cite{WANG20193}, and \mbox{Zhang et al. \cite{dlHARsurvey}}, which offer in-depth discussions on different techniques in HAR research.

Building on this foundation of HAR approaches, our work focuses on representation learning, a key strategy for achieving label efficiency by improving model performance with fewer labeled examples. In the following section, we explore several representation learning paradigms that directly address the challenge of label scarcity in HAR applications.

\subsection{Representation Learning} \label{rl}
We categorize representation learning paradigms based on the degree of supervision used during training: supervised representation learning (Section \ref{srl}), unsupervised representation learning (Section \ref{url}), weakly supervised representation learning (Section~\ref{wsrl}), and self-supervised representation learning (Section \ref{ssrl}). 

In order to illustrate these concepts, let us consider a running example and follow the same notations in this section. Consider a dataset with labels as $D = \{(X, Y)\}$, where $X=(x_{0} \dots x_{i} \dots x_{N})$ represents the input samples, $Y=(y_{0}^{0} \dots y_{i}^{j} \dots y_{N}^{M})$ symbolizes the corresponding labels, $N$ is the number of training samples, $i$ is the sample index, $M$ is the number of classes in the dataset, and $j$ is the class index. In the context of wearable-sensor-based HAR, $X$ would be the readings from different sensors, while $Y$ would be distinct categories of human activities. The objective of representation learning involves training a feature extraction function, denoted 
 as $\textbf{f}(\cdot): \mathbb{R}^{I} \rightarrow \mathbb{R}^{H}$, which maps an input sample $\mathbf{x} \in \mathbb{R}^{I}$ to a feature representation $\textbf{f}(\mathbf{x}) \in \mathbb{R}^{H}$. The core goal is to learn a transformation $\textbf{f}(\cdot)$ that projects raw input data into a new representation space where samples from different classes or categories become more distinct and separable. Ideally, the learned representations $\textbf{f}(\mathbf{x})$ should highlight the discriminative factors that distinguish between distinct classes or concepts, while being invariant to irrelevant variations. By capturing these class-differentiating characteristics in a disentangled and information-preserving manner, the derived representations can significantly improve the performance of models in subsequent tasks such as classification, clustering, regression, or other downstream analyses.

\subsubsection{Supervised Representation Learning} \label{srl}
In supervised representation learning \cite{nozawa2022evaluation}, the training process leverages labeled data, where each input sample $\mathbf{x}$ is associated with a corresponding target label $\mathbf{y}$. The goal is to learn a feature extractor $\textbf{f}(\cdot): \mathbb{R}^{I} \rightarrow \mathbb{R}^{H}$ that maps the raw input $\mathbf{x} \in \mathbb{R}^{I}$ to a representation $\textbf{f}(\mathbf{x}) \in \mathbb{R}^{H}$ that is well suited and effective for the target task. This learned representation $\textbf{f}(\mathbf{x})$ serves as input to a task-specific function $\textbf{g}(\cdot): \mathbb{R}^{H} \rightarrow \mathbb{R}^{M}$, which generates the predicted output $\tilde{\mathbf{y}} = \textbf{g}(\textbf{f}(\mathbf{x})) \in \mathbb{R}^{M}$. The predicted output $\tilde{\mathbf{y}}$ is then compared against the ground truth target $\mathbf{y}$ using a suitable loss function, which guides the optimization of both $\textbf{f}(\cdot)$ and $\textbf{g}(\cdot)$ to learn representations that are discriminative for the supervised task at hand. The training involves minimizing the supervised loss function $\Phi_{srl}$, which could be, for instance, a cross-entropy loss. Following the minimization of the supervised loss $\Phi_{srl}$, the trained $\textbf{f}(\cdot)$ is employed as the feature extraction function to generate the representation. This representation can then be used for various subsequent tasks. 

One of the strengths of supervised representation learning is the acquisition of the feature extraction function $\textbf{f}(\cdot)$ as a direct outcome of supervised training. As an example, models like VGG \cite{simonyan2014very}, initially trained for ImageNet classification, have found extensive utility across diverse visual tasks \cite{girshick2014rich}. A larger dataset in supervised representation learning has been shown to positively impact downstream performance \cite{sun2017revisiting}. However, it is crucial to acknowledge that supervised representation learning heavily depends on available labels, which comes at a significant cost in terms of both time and financial resources. To counter this limitation, unsupervised representation learning has attracted substantial attention within the machine learning and HAR research communities.

\subsubsection{Unsupervised Representation Learning} \label{url} 
Unsupervised representation learning is relevant when labeled data are unavailable. In such a scenario, the objective is to unveil patterns or inherent structure within unlabeled data and to reshape input data into a more insightful representation space without relying on explicit labels. This is usually achieved by training the feature extraction function $\textbf{f}(\cdot)$ through unsupervised tasks, such as dimensionality reduction, data reconstruction, clustering, or generative modeling. The feature extractor $\textbf{f}(\cdot): \mathbb{R}^{I} \rightarrow \mathbb{R}^{H}$ is designed to map the raw input $\mathbf{x} \in \mathbb{R}^{I}$ to a representation $\textbf{f}(\mathbf{x}) \in \mathbb{R}^{H}$ that captures features and structure within the data. This learned representation $\textbf{f}(\mathbf{x})$ then serves as input to downstream tasks. 

One major category of unsupervised representation learning techniques is dimensionality reduction. Linear approaches like principal component analysis (PCA) identify the principal components that explain the maximum variance in the data through mathematical decomposition, allowing for a simplified representation of the original dataset while retaining as much of the original information as possible. Nonlinear dimensionality reduction methods like t-SNE \cite{tsne} or UMAP \cite{mcinnes2020umapuniformmanifoldapproximation} better preserve local neighborhood structures, enabling more accurate representation of complex data manifolds.

Beyond traditional dimensionality reduction, neural-network-based approaches offer powerful alternatives for unsupervised representation learning. Autoencoders learn to compress data into a lower-dimensional latent space and then reconstruct it, forcing the network to capture essential features of the data. Unlike PCA, autoencoders can model complex nonlinear relationships in the data. Various extensions of the basic autoencoder architecture, such as denoising autoencoders that learn robust representations by reconstructing clean inputs from corrupted versions, or variational autoencoders that learn probabilistic latent variable models, further enhance representation quality for downstream~tasks.

Due to its unsupervised nature, unsupervised representation learning can easily be scaled to larger unlabeled datasets at minimal cost. For instance, in the work of \cite{mikolov2017advances}, word representations were trained using an extremely extensive dataset. This characteristic holds significant value, given that the magnitude of the dataset plays a pivotal role in enhancing the real-world performance of models in downstream tasks, as highlighted by studies such as \cite{kaplan2020scaling}.

While unsupervised approaches eliminate the need for labels entirely, they often struggle to capture task-specific representations effectively. This limitation inspired the development of weakly supervised representation learning, which aims to achieve better performance with minimal labeling effort.

\subsubsection{Weakly Supervised Representation Learning} \label{wsrl}
Weakly supervised representation learning methods bridge the gap between fully supervised and unsupervised approaches. By incorporating limited or imperfect labels, these techniques aim to guide representation learning more effectively than unsupervised methods while requiring less complete annotation than fully supervised techniques. Common weakly supervised strategies include multiple instance learning using group-level labels \cite{carbonneau2018multiple}, label propagation from a small labeled subset \cite{iscen2019label}, and contrastive methods with pairwise constraints \cite{le2020contrastive, Sheng2019SMC} (e.g., must-link, cannot-link). The learning process optimizes representations using noisy or incomplete labels, understanding they may be less reliable than full supervision.

In weakly supervised learning, the available labels are incomplete or imprecise. A function $\textbf{g}(\cdot)$ utilizes the learned representations $\textbf{f}(\mathbf{x})$ to predict weak labels $\mathbf{y}_w$, which only provide partial or coarse ground truth for the desired outputs. The predicted weak label, $\tilde{\mathbf{y}}_w=\textbf{g}(\textbf{f}(\mathbf{x})) \in \mathbb{R}^{M}$, will be measured against the true weak label $\mathbf{y}_w$. The loss function $\Phi_{\text{wsrl}}$, which measures the error between predictions and true values, is minimized to optimize the functions $\textbf{f}(\cdot)$ and $\textbf{g}(\cdot)$, thereby learning representations suited for downstream tasks while accounting for imperfect labels. For instance, in contrastive learning with pairwise constraints, the model is guided by information about which data pairs should be similar (must-link) and which should be dissimilar (cannot-link). The learning objective typically involves minimizing the distance between representations of must-link pairs while maximizing the distance between cannot-link pairs. This encourages the model to learn a representation space where semantically similar items are clustered together and dissimilar items are separated, even without knowing the exact class labels.

Despite their effectiveness, weakly supervised methods still require some form of external supervision. Self-supervised representation learning emerged as an alternative that eliminates the need for any external labels by deriving supervisory signals directly from the data itself.

\subsubsection{Self-Supervised Representation Learning} \label{ssrl}
Self-supervised representation learning generates internal supervision from the data through pretext tasks that expose meaningful features. By designing proxy objectives leveraging intrinsic structure within the data, self-supervision transforms the data into their own source of labels.

The general approach involves creating artificial prediction tasks that do not require manual annotations but force the model to learn useful representations. These pretext tasks are designed so that solving them requires understanding important structural aspects of the data. The learning process minimizes the discrepancy between predicted and actual labels derived from these internal supervisory signals.

Numerous pretext tasks have been developed across different domains. In computer vision, these include predicting missing image parts \cite{zhao2021self}, determining relative patch positions, colorizing grayscale images, and predicting image rotations \cite{gidaris2018unsupervised}. For sequential data, common tasks include predicting future steps, sorting scrambled sequences \cite{lee2017unsupervised}, or determining sequence ordering \cite{NEURIPS2021_02e656ad}.

As a concrete example, in the image rotation prediction task \cite{gidaris2018unsupervised}, the authors define a collection of $K$ discrete geometric transformations, denoted as $T = \left\{\textbf{t}(\cdot | r)\right\}_{r=0}^{K}$. Here, $\textbf{t}(\cdot | r)$ represents the operator that applies a geometric transformation labeled $r$ to an image $\mathbf{x}$, resulting in the transformed image $\mathbf{x}^r = \textbf{t}(\mathbf{x} | r)$. The feature extraction function, denoted as $\textbf{f}(\cdot)$, and the rotation prediction function, denoted as $\textbf{g}(\cdot)$, operate on the transformed image $\mathbf{x}^r$ (with the label $r$ unknown to these functions). The predicted label, $\tilde{r} = \textbf{g}(\textbf{f}(\mathbf{x}^r))$, is then compared to the true label $r$. The optimization of $\textbf{f}(\cdot)$ and $\textbf{g}(\cdot)$ is driven by a surrogate supervised loss function, $\Phi_{ssrl}$, which measures the prediction accuracy of the rotation transformations. After training, the learned $\textbf{f}(\cdot)$ is used as the feature extraction function to generate representations for downstream tasks.

These self-supervised methods extend representation learning to scenarios with limited external supervision by generating internal supervisory signals from the data. They offer viable alternatives when external labels are scarce or imperfect. Self-supervised approaches are particularly valuable in domains like HAR with wearables, where collecting labeled data is challenging and expensive.

\subsection{Technical Methodologies}
Building on the review of HAR and deep representation learning methods in \mbox{Sections \ref{har} and \ref{rl}}, we identified and selected the techniques most relevant to this work. These fundamental deep learning architectures and methods have been chosen for their proven effectiveness in capturing temporal patterns and learning discriminative representations from wearable sensor data. Each selected technique addresses specific challenges in HAR: autoencoders \cite{Kramer1991NonlinearPC} for unsupervised representation learning, ResNet \cite{resnet} for handling deep architectures without degradation, Siamese networks \cite{signature1993} for similarity-based learning in weakly supervised settings, temporal convolutional networks (TCNs) \cite{TCN1} for effective temporal pattern extraction, and contrastive loss functions \cite{contrast} for implementing pairwise constraint-based learning.

These methods have been extensively tested and widely adopted in prior studies, providing a strong and reliable foundation for our research. More importantly, they offer the flexibility needed to implement and compare the different representation learning paradigms outlined in Section \ref{rl}. The following subsections present a detailed exploration of each selected technique and how they contribute to our framework for reducing label dependency in HAR.

\subsubsection{Autoencoder} \label{ae}
An autoencoder consists of two primary components: an encoder and a decoder. The encoder serves as the feature extraction function, $\textbf{f}(\cdot): \mathbb{R}^{I} \rightarrow \mathbb{R}^{H}$, while the decoder performs the transformation $\textbf{g}(\cdot): \mathbb{R}^{H} \rightarrow \mathbb{R}^{I}$, aiming to reconstruct the original input $\mathbf{x}$ from its representation $\textbf{f}(\mathbf{x})$. Formally, this process is defined as $\tilde{\mathbf{x}} = \textbf{g}(\textbf{f}(\mathbf{x})) \in \mathbb{R}^{I}$.

The optimization of $\textbf{f}(\cdot)$ and $\textbf{g}(\cdot)$ is guided by the data reconstruction loss:
\begin{align} 
\Phi_{ae}(\mathbf{x}) = \frac{1}{N} \sum_{i=0}^{N} ||\mathbf{x}_{i} - \textbf{g}(\textbf{f}(\mathbf{x}_{i}))||^2 
\end{align}
here, 
 $||\cdot||$ denotes a metric such as the Euclidean distance. This loss function minimizes the difference between the original input $\mathbf{x}_i$ and its reconstruction $\tilde{\mathbf{x}}_i = \textbf{g}(\textbf{f}(\mathbf{x}_i))$, encouraging $\tilde{\mathbf{x}}_i$ to closely approximate $\mathbf{x}_i$.

A high-quality reconstruction implies that the representation $\textbf{f}(\mathbf{x}_i)$ effectively preserves the essential information from $\mathbf{x}_i$. Consequently, this representation becomes valuable for downstream tasks such as classification or clustering.

Autoencoders have been successfully applied across various domains. For instance, Vincent et al. \cite{vincent2008extracting} used denoising autoencoders for robust feature learning, Kingma and Welling \cite{kingma2022autoencodingvariationalbayes} introduced variational autoencoders for generative modeling, and Sakurada and Yairi \cite{10.1145/2689746.2689747} employed autoencoders for anomaly detection in time-series data.

\subsubsection{ResNet}
\begin{figure}[H]
\includegraphics[width=0.85\textwidth]{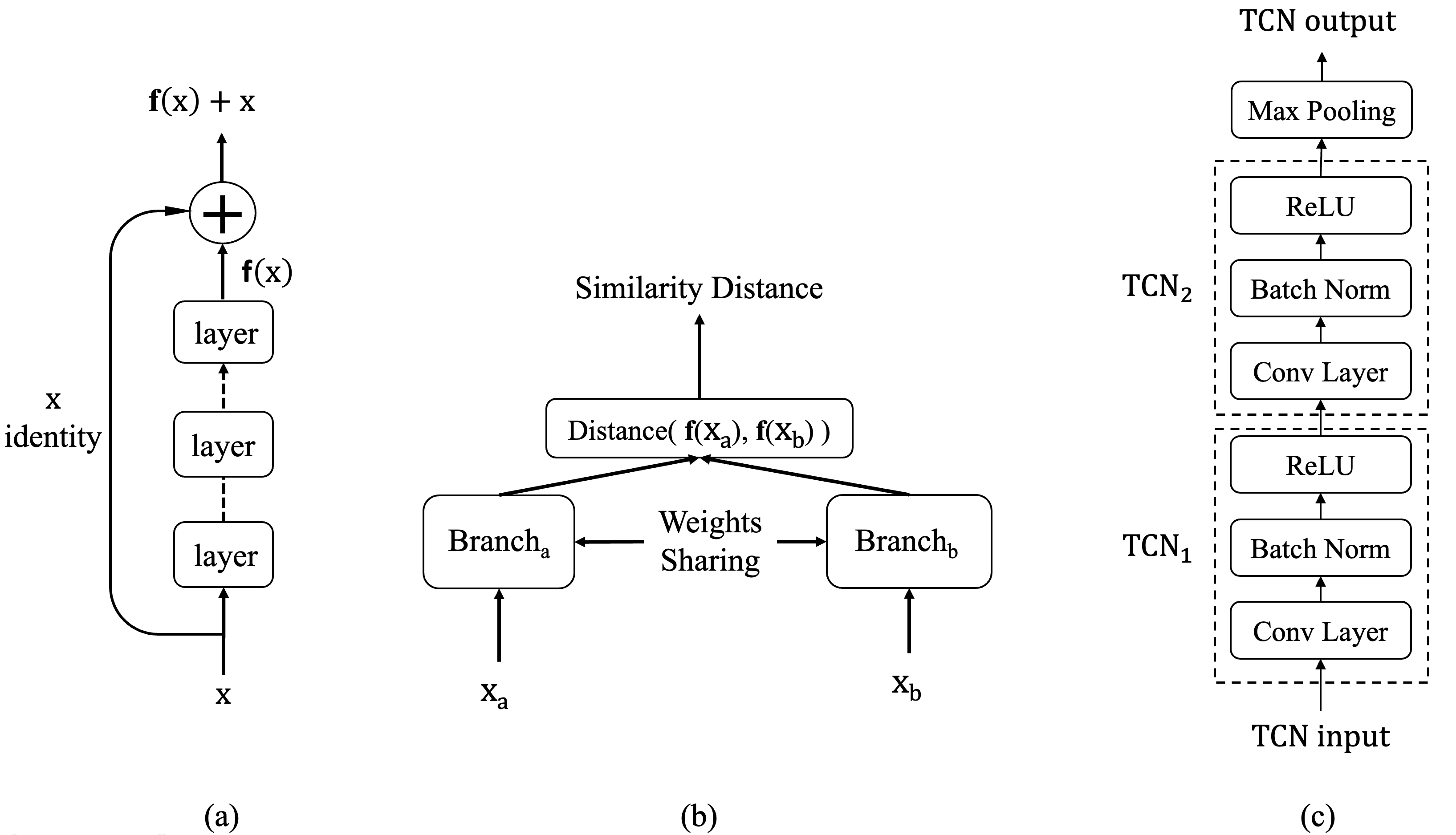} 
\caption{(\textbf{a}) ResNet, 
 (\textbf{b}) Siamese architecture, and (\textbf{c}) TCN block.} 
\label{resSiamese}
\end{figure}

The ResNet architecture provides an effective means to increase network capacity while addressing the degradation of feature learning commonly observed in standard deep neural networks (DNNs). It achieves this through skip connections or shortcuts, as illustrated in Figure~\ref{resSiamese}a for a single ResNet block. These connections allow certain layers to be bypassed, ensuring that information flows more efficiently through the network.

The output of a ResNet block is computed by summing the result of the last layer with the input to the block. When the input and output tensors differ in shape, a linear layer of matching size is used for the residual connection. If the tensors share the same shape, an identity function is employed instead. This design ensures seamless integration of residual connections, enhancing learning stability and improving overall network performance.

ResNet architectures have demonstrated remarkable success in numerous applications. He et al. \cite{resnet} showed their superior performance in image recognition tasks, \mbox{Kolesnikov et al. \cite{DBLP:journals/corr/abs-1901-09005}} adapted ResNets for self-supervised visual representation learning, and Maweu et al. \cite{9421374} applied them to time-series data to improve diagnostic accuracy in healthcare.

\subsubsection{Siamese Neural Networks}
A Siamese neural network consists of two branches with shared weights, as depicted in Figure~\ref{resSiamese}b. This architecture processes two distinct inputs, generating comparable representation vectors that encapsulate each input's unique features. These vectors are then fed to a metric layer \cite{sentenceSiamese} or a learned metric network, which measures the similarity between the two inputs.

Siamese networks are widely applied in nonlinear metric learning across domains such as computer vision and speech recognition. For instance, Siamese CNNs have been used to develop sophisticated similarity metrics for face verification \cite{faceverification}. In natural language processing, Mueller and Thyagarajan \cite{sentenceSiamese} introduced a Siamese RNN to evaluate semantic similarity between sentences, while Neculoiu et al. \cite{textSiamese} employed one for job title normalization in recruitment analysis. Zeghidour et al. \cite{jointlearning} leveraged a multi-output Siamese network for speaker and phonetic similarity detection, demonstrating the versatility of this architecture.

\subsubsection{Temporal Convolutional Networks}
Temporal convolutional networks (TCNs) serve as a foundational component in the proposed approaches. A variation of the TCN block, illustrated in Figure~\ref{resSiamese}c, uses convolutional layers to extract local patterns from input sequences while enabling translational invariance for these patterns across the data sequence.

Each convolutional layer is followed by batch normalization (BN) \cite{BNA}, which stabilizes optimization by reparameterizing the problem into a more tractable form. BN accelerates training and provides regularization, preventing early saturation of the nonlinear activation functions. The rectified linear unit (ReLU) \cite{RELU} serves as the activation function, further enhancing nonlinear representation learning.

To reduce temporal dimensionality and introduce a degree of translational invariance, a temporal max-pooling layer is added after every two TCN blocks. Max-pooling selects the maximum value within predefined regions, achieving a subsampling effect.

TCNs have demonstrated efficacy in capturing temporal dependencies across diverse applications. For example, Lea et al. \cite{TCN2} utilized a TCN for action segmentation and detection, Zhang et al. \cite{charTCN} applied it for text classification, and Bednarski et al. \cite{bednarski2022temporal} employed it in health informatics for predicting clinical length of stay and mortality.

\subsubsection{Contrastive Loss} \label{lossfunc235}
In our experiments, we adopt the contrastive loss function to train the model. This function operates on triplets $(\mathbf{x}_a, \mathbf{x}_b, \mathbf{y})$, where $\mathbf{x}_a$ and $\mathbf{x}_b$ are input samples and $\mathbf{y} \in \{0,1\}$ is a binary label indicating whether the samples are similar ($\mathbf{y}=1$) or dissimilar ($\mathbf{y}=0$). The model generates activity representations $\textbf{f}(\mathbf{x}_a)$ and $\textbf{f}(\mathbf{x}_b)$. The contrastive loss evaluates the similarity distance $Dist_s$ between these representations using the Euclidean distance metric, defined as:
\begin{align} 
Dist_s(\mathbf{x}_a, \mathbf{x}_b) = \|\textbf{f}(\mathbf{x}_a) - \textbf{f}(\mathbf{x}_b)\|_2  
\label{dists}
\end{align}

This distance measures the similarity or dissimilarity between inputs in the representation space. The loss function is designed to minimize the distance for similar pairs while maximizing it for dissimilar pairs, enabling the model to capture meaningful patterns for activity discrimination.

For simplicity, the similarity distance $Dist_s(\mathbf{x}_a, \mathbf{x}_b)$ is denoted as $D$. Using this notation, the loss for each training sample is expressed as:
\begin{align} 
\Phi_{simi_i}(\mathbf{x}_{a_i}, \mathbf{x}_{b_i}, \mathbf{y}_i) = \mathbf{y}_i L_s(\mathbf{x}_{a_i}, \mathbf{x}_{b_i}) + (1 - \mathbf{y}_i) L_d(\mathbf{x}_{a_i}, \mathbf{x}_{b_i})  
\label{distsEqu1}
\end{align}

The overall loss for all training samples is then given by:
\begin{align} 
\Phi_{simi}(\mathbf{x}_a, \mathbf{x}_b, \mathbf{y}) = \sum_{i=0}^{N} \Phi_{simi_i}(\mathbf{x}_{a_i}, \mathbf{x}_{b_i}, \mathbf{y}_i)  
\label{distsEqu2}
\end{align}
here, $(\mathbf{x}_{a_i}, \mathbf{x}_{b_i}, \mathbf{y}_i)$ represents the $i$-th training sample, and $N$ is the total number of samples. $L_s$ and $L_d$ denote the loss terms for positive (similar) pairs $(\mathbf{y}=1)$ and negative (dissimilar) pairs $(\mathbf{y}=0)$, respectively. These loss terms are defined as:
\begin{align} 
&L_d = \frac{1}{2} \big(\max(0, \delta - D)\big)^2  \label{ld}  \\ 
&L_s = \frac{1}{2} D^2  \label{ls}  
\end{align}

The hyperparameter $\delta$, known as the margin, ensures that only negative pairs with a similarity distance smaller than $\delta$ contribute to the loss. This mechanism focuses the learning process on difficult negative samples near the decision boundary while ignoring well-separated negative pairs.

\textls[-15]{Contrastive loss functions have been widely adopted in various domains. \mbox{Chen et al. \cite{DBLP:journals/corr/abs-2002-05709}}} employed contrastive learning for visual representations, Xu et al. \cite{10.1145/3593590} used similar techniques for learning sentence representations, and Pan et al. \cite{pan-etal-2021-contrastive} applied contrastive loss for multilingual neural machine translation.

\subsection{Positioning Within Current Research}
{Recent developments in wearable-based HAR have demonstrated the potential of deep learning architectures and various learning paradigms to improve performance and address labeling challenges. A wide range of approaches have been explored, covering supervised, unsupervised, weakly supervised, and self-supervised learning strategies, each with its own strengths and trade-offs.

From an architectural perspective, several state-of-the-art models have been proposed for time-series representation learning. Transformer-based approaches such as PatchTST}~\cite{Yuqietal-2023-PatchTST} {have shown strong performance in capturing long-range temporal dependencies and are increasingly applied in HAR tasks. Likewise, InceptionTime} \cite{InceptionTime} {and TimesNet} \cite{TimesNet} {represent powerful architectures that extract multi-scale temporal features through hierarchical design. In our work, we adopt TCN and ResNet backbones for their proven effectiveness and computational efficiency on sensor data, while noting that these newer architectures present promising directions for future exploration.

In terms of learning paradigms, recent self-supervised HAR studies offer valuable insights into pretraining strategies. Haresamudram et al.} \cite{10.1145/3550299} {conduct an extensive evaluation of self-supervised methods across multiple HAR settings. Yuan et al.} \cite{Yuan_2024} {leverage large-scale wearable data to train generalizable representations using self-supervision. Cheng et al.} \cite{10.1016/j.knosys.2023.110789} {propose contrastive learning techniques that emphasize invariance to augmentation, and Qian et al.} \cite{10.1145/3534678.3539134} {examine how contrastive learning can be adapted to small-scale HAR datasets.

Our study contributes to this growing body of work by comparing the effectiveness of different learning paradigms, supervised, unsupervised, weakly supervised, and self-supervised, and exploring how they can be combined. In particular, we investigate the integration of domain-informed self-supervision with weak supervision using pairwise constraints. This perspective provides a systematic view of the label--efficiency trade-offs inherent to each approach and highlights potential synergies between them for sensor-based~HAR.

By situating our experiments within this broader landscape, we aim to offer practical guidance on designing HAR systems under various labeling constraints and to encourage further exploration of hybrid strategies in representation learning.}

\section{Approaches} 
In this section, we explore strategies for reducing reliance on labeled data in HAR systems through deep representation learning. Our discussion covers supervised, unsupervised, weakly supervised, self-supervised, and multi-task learning paradigms. Existing HAR models primarily rely on supervised learning for high accuracy but require extensive labeled data, making them costly and time consuming. In contrast, unsupervised methods reduce labeling needs but often perform worse due to the lack of explicit supervision. This fundamental trade-off between label efficiency and model effectiveness motivates our exploration of alternative learning paradigms.

By adapting these approaches to the intrinsic characteristics of human activities, we enhance their applicability to wearable-sensor-based HAR. Leveraging diverse learning paradigms and domain knowledge enables us to minimize labeling requirements while supporting the acquisition of effective representations for sensor-driven HAR tasks. Our primary goal is to systematically compare and investigate these paradigms and reduce the need for large labeled datasets, facilitating the deployment of deep-learning-based HAR in real-world scenarios.

\subsection{Supervised Learning Approach}
Supervised deep learning techniques have seen development and utilization within HAR systems \cite{cnnHAR,dnnHARbenchmark,cnnLSTMHAR}. These techniques have the capacity to automatically extract features from data and have proven valuable in HAR applications. Nonetheless, they still demand explicit labels to supervise model training. In real-world HAR use cases, manually labeling extensive sensor datasets is often infeasible, limiting the scalability of purely supervised methods. In our experiments, we use a TCN as the baseline model to evaluate the effectiveness of supervised learning in HAR.

\subsubsection*{Model Architecture and Training Process}

\begin{figure}[H]

\includegraphics[width=0.95\textwidth]{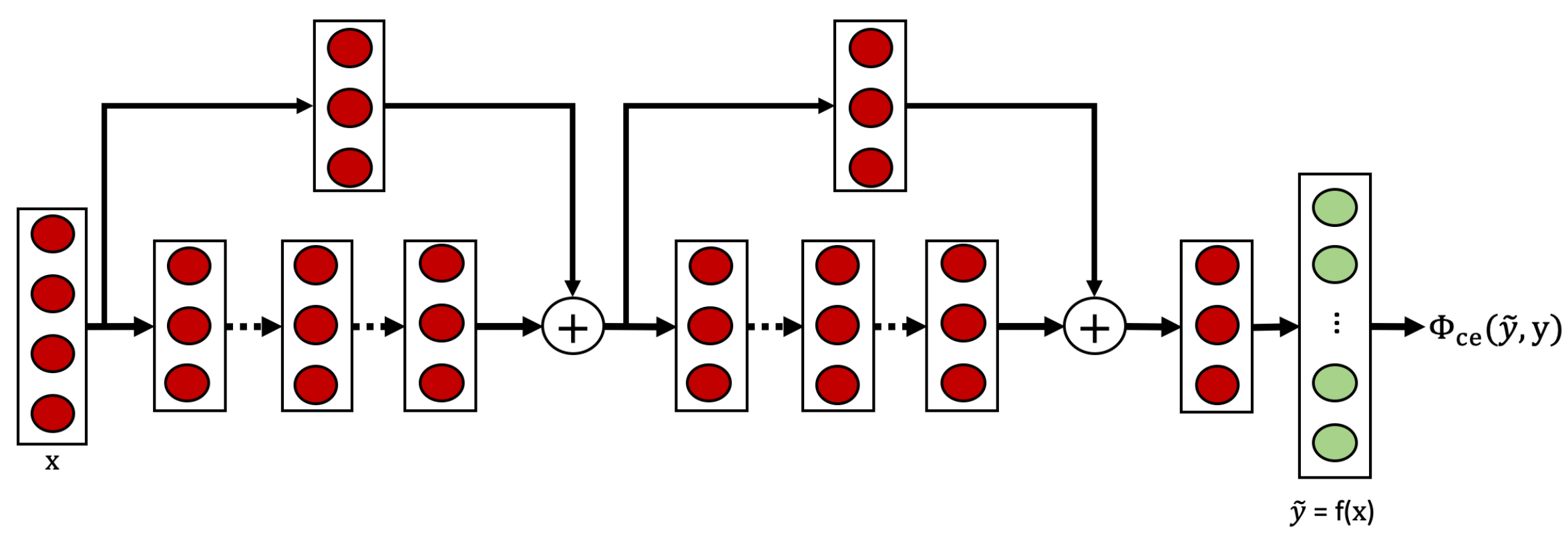} 
\caption{The TCN architecture used in the supervised approach.} \label{SLArch}

\end{figure}

The supervised baseline employs a TCN architecture as illustrated in Figure~\ref{SLArch}. This architecture incorporates ResNet's skip connections (shown in Figure~\ref{resSiamese}a) within the TCN blocks (shown in Figure~\ref{resSiamese}c), creating a residual temporal convolutional network that benefits from both techniques. The model processes raw sensor data through a series of TCN blocks, each containing dilated causal convolutions that capture temporal patterns at increasing time scales. This architecture enables the network to effectively model long-range dependencies while maintaining computational efficiency.

\vspace{-6pt}

Each TCN block contains a sequence of 1D convolutional layers, with each convolutional layer followed by batch normalization and ReLU activation. Residual connections are implemented within blocks to facilitate gradient flow during training. The model concludes with a max pooling layer and a fully connected layer that maps to the target activity classes.

The supervised model is trained using cross-entropy loss $\Phi_{ce}(\mathbf{x}, \mathbf{y})$, which measures the discrepancy between predicted class probabilities and the ground truth labels:
\begin{align}
\Phi_{ce}(\mathbf{x}, \mathbf{y}) = -\sum_{j=0}^{M} \mathbf{y}_j \log(\tilde{\mathbf{y}}_j)
\end{align}
where $\mathbf{y}$ is the ground truth labels, $\tilde{\mathbf{y}} = \textbf{f}(\mathbf{x})$ represents the predicted class probabilities, and $M$ is the number of activity classes. This loss function encourages the model to learn discriminative representations that map directly to activity categories, but requires complete activity labels for all training samples.

\subsection{Unsupervised Learning Approach} \label{ula}
To mitigate the reliance on labeled data, researchers have explored unsupervised learning techniques such as clustering and autoencoder-based feature learning \cite{Sheng2020UnsupervisedEL}. These methods extract latent representations from sensor data without requiring explicit activity labels. While this reduces annotation costs, the performance of unsupervised models often lags behind their supervised counterparts, as they lack guidance on what features are most relevant for distinguishing activities. For our unsupervised baseline, we adopt an autoencoder to learn meaningful activity representations from unlabeled sensor data.

Unlike the supervised approach, which processes raw sensor data directly, our unsupervised method operates on handcrafted features extracted from the sensor data. This design choice is deliberate and addresses several challenges inherent to unsupervised learning for HAR. Without the guidance of labels to direct feature learning, autoencoders applied directly to raw wearable sensor data can struggle to identify meaningful activity patterns among sensor noise and irrelevant variations. Handcrafted features incorporate domain expertise about which signal characteristics are most relevant for distinguishing human activities, providing valuable inductive bias that compensates for the absence of explicit supervision. Additionally, these statistical features help reduce dimensionality while preserving essential activity information, enabling the autoencoder to focus on learning higher-level activity representations rather than basic signal processing. Using handcrafted features is particularly valuable in the unsupervised context, where the model must discover meaningful structure without label guidance.

\subsubsection{Feature Extraction} \label{featExtract}
This section focuses on the handcrafted features used in the experiments. In the feature extraction stage, the segmented raw sensor signals are converted into feature vectors. Formally, let $\mathbf{r}_i$ denote the sample $i$ in the set of the segmented raw sensor signals, $\mathbf{x}_i$ the converted feature vector, and $C$ the feature extraction function. Then, the feature extraction can be defined as Equation \eqref{featconv}:
\begin{align}
\mathbf{x}_i = C(\mathbf{r}_i) \label{featconv}
\end{align}

Finally, $\mathbf{x}_i$ is used as the input to the model. Table \ref{statisticalFeat} illustrates the statistical high-level features that are used in this approach. Mean, variance, standard deviation, and median, which are the most commonly adopted features in HAR research works, are used in this approach. In addition, some other features, which have been shown to be efficient in previous works~\cite{transitionAware}, are included here as well. For example, the feature interquartile range ($iqr$). Quartiles ($Q_1$, $Q_2$, and $Q_3$) divide the time series signal into quarters. Using this, $iqr$ is the measure of variability between the upper and lower quartiles, $iqr = Q_3 - Q_1$. 
\begin{table}[H]
\centering
\caption{List 
 of the used statistical features. Note: $W$ denotes the window length of the signal.}\label{statisticalFeat}\smallskip
\scalebox{1}{%
\begin{tabularx}{\textwidth}{CC}
\toprule
\textbf{Feature Extraction Function} &  \textbf{Description} \\
\midrule
$mean(r_i) = \frac{1}{W}\sum_{k=0}^{W} r_{i_k} $ & Mean \\
$var(r_i) = \frac{1}{W}\sum_{k=0}^{W} (r_{i_k} - mean(r_i))^2 $ & Variance \\
$std(r_i) = \sqrt{var(r_i)} $  & Standard deviation \\
$median(r_i)$ & Median values\\
$max(r_i)$ & Largest values in array \\
$min(r_i)$ & Smallest value in array \\
$iqr(r_i) = Q_{3}(r_i) - Q_{1}(r_i) $ & Interquartile range \\
\bottomrule
\end{tabularx}}
\end{table}

The aforementioned features are computed separately for each axis of the sensor data. Since the data from different sensors are time-synchronized, it is possible to combine features derived from multiple sensor modalities. During the training process, the autoencoder takes the derived features as input and learns to retain the key information while discarding the unnecessary reconstruction-irrelevant components. The encoder effectively projects the feature representations into a lower-dimensional subspace that captures the most informative characteristics for the reconstruction task.

\subsubsection{Model Architecture and Training Process}
\begin{figure}[H]

\includegraphics[width=0.7\textwidth]{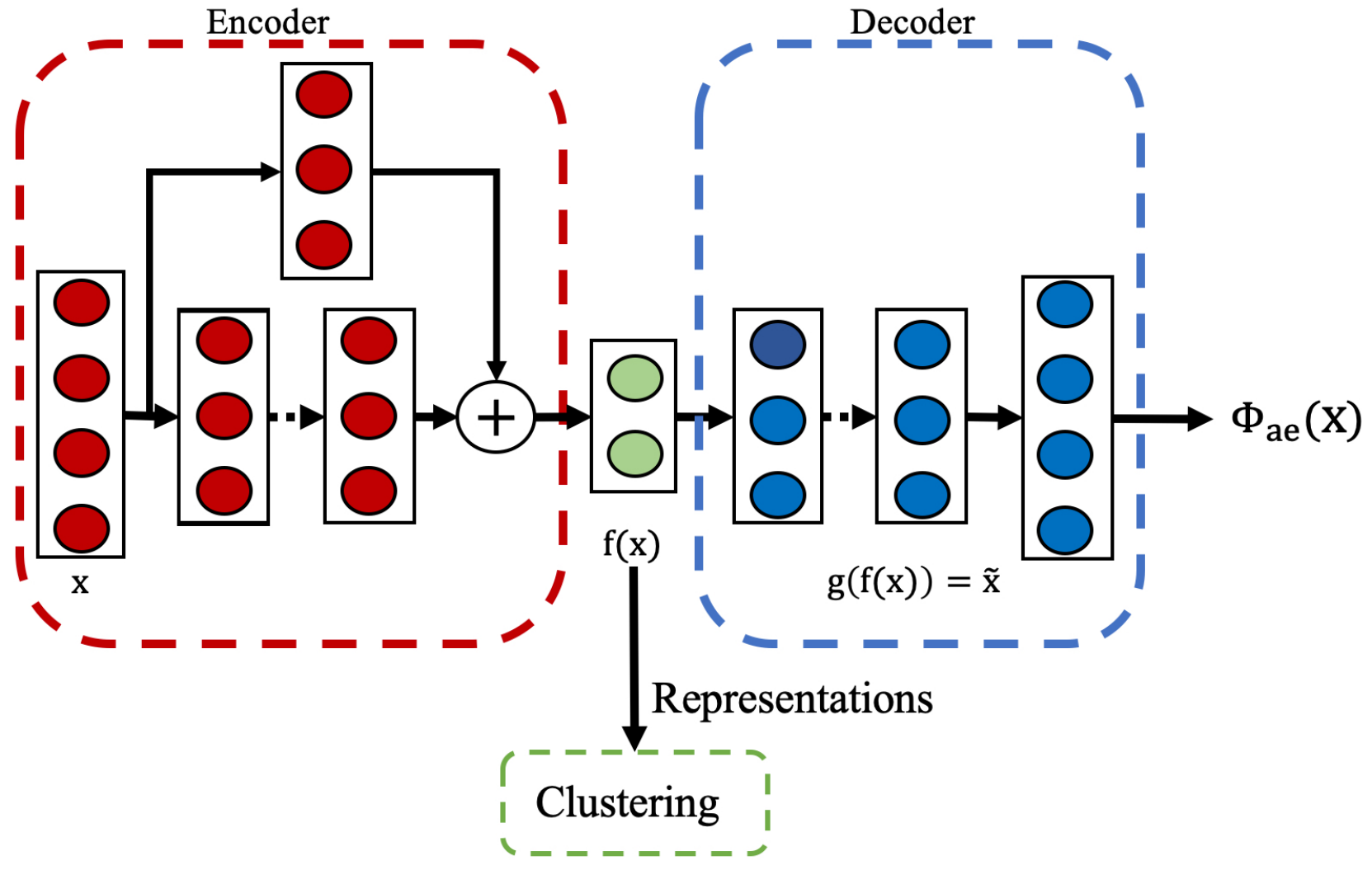} 
\caption{The residual autoencoder architecture used in the unsupervised approach.} \label{ULArch}

\end{figure}

Our unsupervised approach utilizes an autoencoder architecture as shown in Figure~\ref{ULArch}. The model consists of an encoder and a decoder network. The encoder maps the input data to a lower-dimensional latent representation, while the decoder attempts to reconstruct the original input from this compressed representation.

The encoder employs a residual-based architecture composed of residual blocks with progressively decreasing dimensions, followed by batch normalization and ReLU activation. The residual connections help preserve important features and improve the gradient flow during training. The final encoder layer outputs a condensed latent representation, $\mathbf{f}(\mathbf{x})$, which captures the fundamental characteristics of the input. The decoder adopts a simpler structure, consisting of fully connected layers with progressively increasing dimensions, and concludes with an output layer that matches the input dimensions, as the reconstruction task does not require deep residual connections.


{The model is trained using the reconstruction loss already defined in Section}~\ref{ae}. {This loss function encourages the model to learn a compact representation that preserves the important properties of the input data without requiring any activity labels.} After training, the encoder part of the neural network serves as a feature extractor, generating representations that capture the intrinsic structure of the input data for downstream clustering.

While the unsupervised approach reduces the reliance on labeled data, it still faces challenges in learning discriminative representations specifically relevant to activity recognition tasks. To address this limitation while still minimizing labeling requirements, we next explore a weakly supervised paradigm that leverages minimal supervision signals in the form of pairwise constraints.

\subsection{Weakly Supervised Single-Task Approach} \label{weaksup}
To address the challenge of minimizing the need for extensive supervision during model training, we introduce an initial weakly supervised single-task approach using siamese networks to identify human activities within sensor data streams. This method trains a Siamese network to provide a similarity metric, enabling activity clustering without the strict requirement of explicitly labeled data.

\subsubsection*{Model Architecture and Training Process}
\begin{figure}[H]

\includegraphics[width=0.8\textwidth]{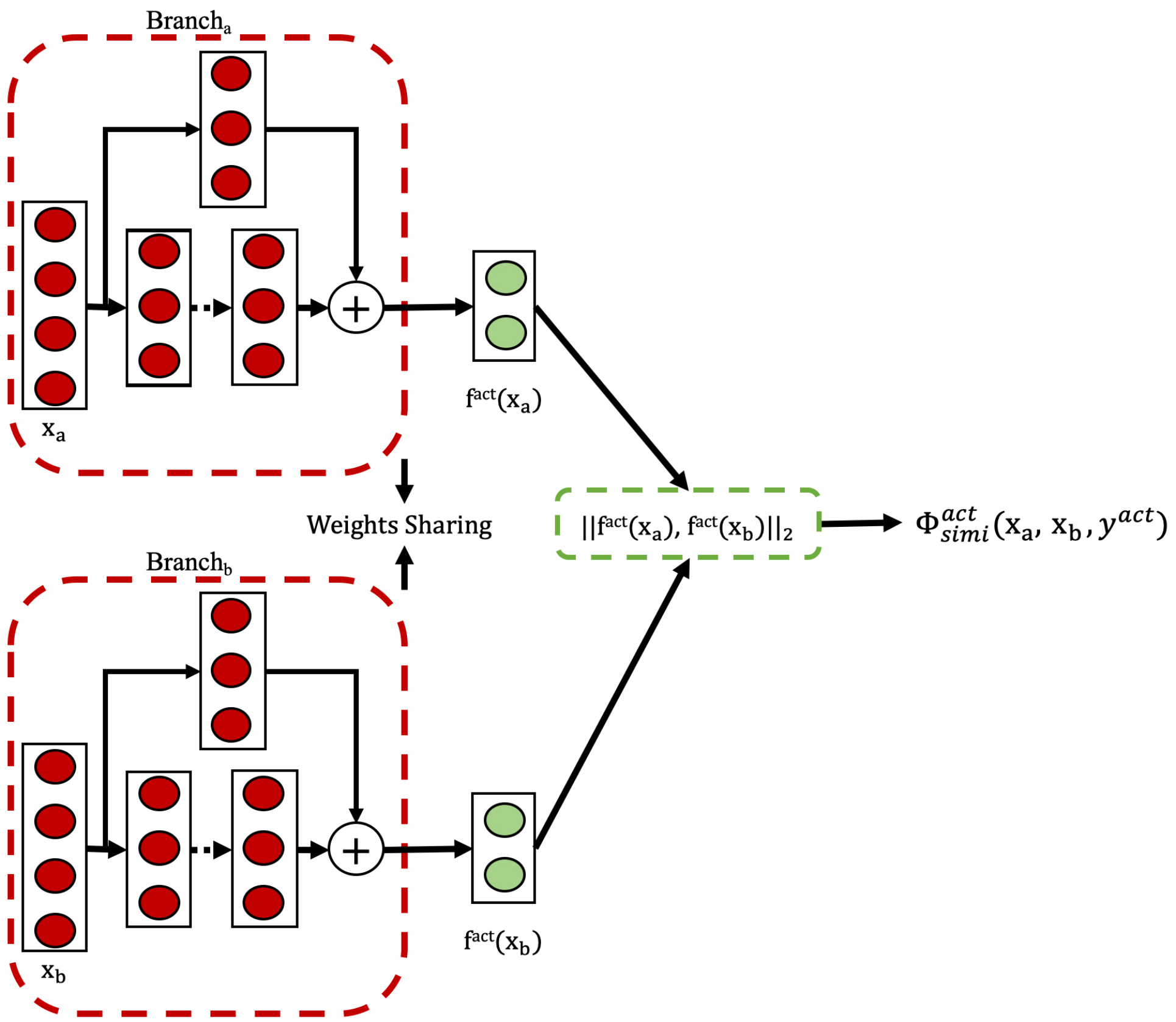} 
\caption{The Siamese architecture used in the weakly supervised single-task approach.} \label{siameseArch}

\end{figure}

Our model uses a Siamese architecture, as depicted in Figure~\ref{siameseArch}. For any pair of input sequences representing human activity data, denoted as $(\mathbf{x}_{a}, \mathbf{x}_{b})$, the Siamese network learns to map them to a shared representation space $(\textbf{f}^{act}(\mathbf{x}_{a})$, $\textbf{f}^{act}(\mathbf{x}_{b}))$ $\in \mathbb{R}^{H}$. The layers following this dual-branch architecture constitute a similarity function, which measures the distance between the two extracted representation vectors. Our Siamese networks maintain weight sharing across their twin branches. Each branch employs identical TCN blocks. Outputs from both branches are subsequently processed using fully connected (FC) layers.

The model is trained using triplets $(\mathbf{x}_a, \mathbf{x}_b, \mathbf{y}^{act})$, where $\mathbf{x}_a$ and $\mathbf{x}_b$ are segmented activity sequences and $\mathbf{y}^{act} \in \left\{0, 1\right\}$ indicates if they are of the same $(\mathbf{y}^{act}=1)$ or different $(\mathbf{y}^{act}=0)$ activity type. 

{We adopt pairwise similarity constraints to supervise representation learning. This choice provides a clean and interpretable formulation that aligns with our aim of isolating the effects of different learning paradigms under consistent architectural and supervision settings. While other formulations such as triplet-based or relative constraints are possible, pairwise supervision offers a simple and effective baseline for evaluating the benefits of weak supervision in activity recognition.}

{As defined in Section}~\ref{lossfunc235}, {we use a contrastive loss function to learn a mapping} $\textbf{f}^{act}(\mathbf{\cdot})$ {that captures critical similarities between the input pairs. If} $\mathbf{y}^{act}=1$, {the representations} $\textbf{f}^{act}(\mathbf{x}_a)$ {and} $\textbf{f}^{act}(\mathbf{x}_b)$ {should be embedded closer together. If} $\mathbf{y}^{act}=0$, {the representations should be farther apart. The activity-specific contrastive loss is defined as:}
\begin{align} 
\Phi_{simi}^{act}(\mathbf{x}_a, \mathbf{x}_b, \mathbf{y}^{act}) = \sum_{i=1}^{N} \Phi_{simi_i}^{act}(\mathbf{x}_{a_i}, \mathbf{x}_{b_i}, \mathbf{y}_{i}^{act})  
\label{distsEqu3}
\end{align}
where $\mathbf{x}_{a_i}$, $\mathbf{x}_{b_i}$, and $\mathbf{y}_{i}^{act}$ represent individual training samples, and $N$ is the total number of such samples. The variables $\mathbf{x}_a$, $\mathbf{x}_b$, and $\mathbf{y}^{act}$ on the left side of the equation denote the entire collections of samples.

\subsection{Weakly Supervised Multi-Task Approach} \label{weakup_multitask}
In machine learning, models are typically trained to optimize a single metric by specializing on one particular task. However, this narrow focus can neglect useful information from related tasks that could improve performance on the original metric of interest. Multi-task learning provides a technique to harness these additional signals by sharing representations across related tasks. Rather than learning in isolation, the shared representation is trained concurrently on multiple objectives. This enables the model to learn generalized patterns that transfer and benefit all tasks. By leveraging inter-task relationships, multi-task learning can enhance model performance on the original target metric beyond single-task specialized approaches. The joint training process allows complementary signals from related tasks to regularize and inform the shared representation. Motivated by this observation, we extended the weakly supervised single-task approach in Section~\ref{weaksup} to include one more related task, person identification, and conducted experiments to evaluate the effectiveness of the weakly supervised multi-task approach on HAR and person identification tasks.

\subsubsection*{Model Architecture and Training Process}

To achieve these capabilities, we adopted a model architecture as illustrated in Figure~\ref{triam}. {This architecture extends the single-task Siamese network from Section}~\ref{weaksup} {by adding a second output head for person identification alongside the original activity recognition branch. Two fully connected layers are connected to the TCN networks and are responsible for processing the activity representations} $(\textbf{f}^{act}(\mathbf{x}_a), \textbf{f}^{act}(\mathbf{x}_b))$ {and the person representations} $(\textbf{f}^{pers}(\mathbf{x}_a), \textbf{f}^{pers}(\mathbf{x}_b))$, {respectively. The weights of the FC layers are shared within each representation learning task.
\vspace{-6pt}

\begin{figure}[H]
\begin{center}
\includegraphics[width=1.0\textwidth]{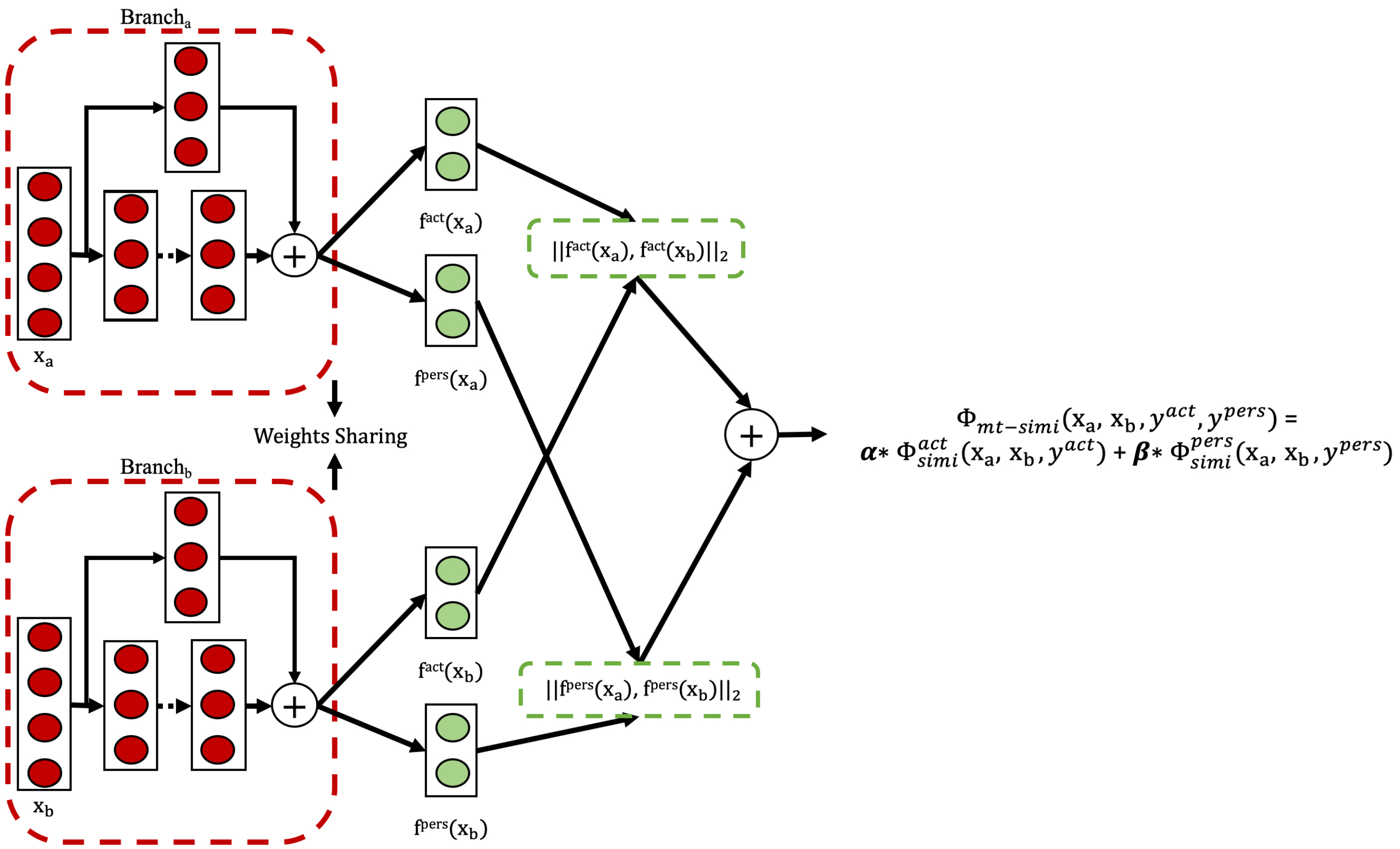} 
\caption{The extended Siamese architecture used in the weakly supervised multi-task approach.} \label{triam}
\end{center}
\end{figure}

\vspace{-12pt}

This study adopts a dual-task setup involving activity recognition and person identification. These two tasks were chosen because they are both semantically meaningful in the context of wearable-sensor-based HAR and offer complementary signals for representation learning. While the multi-task framework can be extended to additional tasks, our focus in this work remains on these two to maintain clarity and tractability in evaluation.}

{Formally, consider a quadruple} $(\mathbf{x}_a, \mathbf{x}_b, \mathbf{y}^{act}, \mathbf{y}^{pers})$, {where} $\mathbf{x}_a$ and $\mathbf{x}_b$ {are the input data sequences (e.g., sensor readings), and} $\mathbf{y}^{act}, \mathbf{y}^{pers} \in \{0, 1\}$ {denote the semantic relationships between the input pair in terms of activity and person, respectively.} Here, $\mathbf{y}^{act}=0$ or $\mathbf{y}^{pers}=0$ {indicates that} ${\mathbf{x}_a, \mathbf{x}_b}$ {is a semantically dissimilar pair, i.e., they correspond to different activities or are performed by different persons. Conversely,} $\mathbf{y}^{act}=1$ or $\mathbf{y}^{pers}=1$ {indicates a semantically similar pair.}

{Following the contrastive loss framework defined in Section}~\ref{lossfunc235}, {the model learns two mappings}, $\textbf{f}^{act}(\mathbf{x})$, and $\textbf{f}^{pers}(\mathbf{x})$ {that encode the relevant semantic relationships between input pairs} $(\mathbf{x}_a, \mathbf{x}_b)$ {in their respective representation spaces. If} $\mathbf{y}^{act}=1$ (or $\mathbf{y}^{pers}=1$){, the representations} $\textbf{f}^{act}(\mathbf{x}_a)$ and $\textbf{f}^{act}(\mathbf{x}_b)$ (or $\textbf{f}^{pers}(\mathbf{x}_a)$ and $\textbf{f}^{pers}(\mathbf{x}_b)$) {should be mapped to nearby positions. If} $\mathbf{y}^{act}=0$ (or $\mathbf{y}^{pers}=0$), {they should be pushed apart.}

{The loss function is applied separately in each representation space. The final multi-task loss is defined as a weighted sum, with} $\alpha$ and $\beta$ {as the corresponding weights for each task, as shown in Equation} \eqref{mtloss}.
\begin{align}
\begin{split}
&\Phi_{mt-simi}(\mathbf{x}_a, \mathbf{x}_b, \mathbf{y}^{act}, \mathbf{y}^{pers})=\\
&\alpha \cdot \Phi_{simi}^{act}(\mathbf{x}_a, \mathbf{x}_b, \mathbf{y}^{act})+ \beta \cdot \Phi_{simi}^{pers}(\mathbf{x}_a, \mathbf{x}_b, \mathbf{y}^{pers}) \label{mtloss}
\end{split}
\end{align}
here, $\Phi_{simi}^{act}$ and $\Phi_{simi}^{pers}$ represent the contrastive loss functions for activity and person similarity, respectively.

\subsection{Self-Supervised Approach} \label{ssa}
As discussed previously in Sections \ref{ae} and \ref{ula}, autoencoders learn to ensure that the reconstructed output closely resembles the initial input. When a high-quality reconstruction can be generated from the encoded representation, it indicates that the representation preserves essential information from the input data, making it potentially useful for downstream tasks like classification and clustering.

However, simply memorizing information for reconstruction is often inadequate for learning useful representations. Standard reconstruction objectives inevitably capture noise and minute details present in the input data, which may be irrelevant or even detrimental for subsequent tasks. To enhance the utility of representations for specific tasks, the autoencoder training objective should incorporate tailored guidance that prioritizes capturing relevant information while filtering out noise and extraneous details.

To address these limitations, we propose a self-supervised approach that leverages two key relationships in human activity data: the temporal consistency of time series and feature consistency in the feature space. Similar to our unsupervised approach in Section~\ref{ula}, this method operates on the handcrafted statistical features described in \mbox{Section~\ref{featExtract}}, ensuring consistent comparison between approaches while incorporating domain knowledge necessary for effective representation learning.

Our self-supervised approach is built upon specialized consistency objectives that guide the representation learning process toward capturing activity-relevant patterns. In the following subsections, we first describe these two consistency objectives that form the core of our approach, followed by the model architecture that integrates these components into a unified framework.

\subsubsection{Temporal Consistency}
Intuitively, a human activity can be decomposed into two components, a temporally varying component and a temporally stationary component. Specifically, certain dynamic properties of a single activity can vary over time. For example, while walking, the body pose varies over time: left foot and right foot alternately step forward. This type of dynamic property is recorded in the sensor data, too, and we refer to it here as the temporally varying component. On the other hand, no matter how the body pose varies over time, the semantic content of the activity remains the same. Namely, left foot and right foot can step forward alternately, but the type of the activity is still walking. We refer to this part as the temporally stationary component. 

Based on this quality of human activity, the temporal consistency loss forces temporally close data samples to be similar to one another and ignores the difference in the temporally varying component. It is motivated by the intention that the semantic content (i.e., the type of the activity in which we are interested) should vary relatively infrequently over time. If the data samples are temporally close to each other, they may represent the same type of activity, even as they may be very distant in terms of the Euclidean distance in the sensor data space. The temporal consistency loss preserves the temporal continuity of the sensor data.

More formally, let $\mathbf{x}_{i^{t}}$ denote data sample $i$, which occurs at time $t$ during the course of an activity. For each sample, we define $P_{i^{t}}$ as the set of indices of its $|P|$ temporal neighbors, which are samples that occur close in time to $\mathbf{x}_{i^{t}}$. The temporal consistency loss $\Phi_{tc}$ for $\mathbf{x}_{i^{t}}$ is then defined as:
\begin{align}
\Phi_{tc}(\mathbf{x}_{i^{t}}) = \frac{1}{|P|}\sum_{p \in P_{i^{t}}}||\mathbf{x}_{p} - \tilde{\mathbf{x}_{i^{t}}}||^{2} \label{losstc}
\end{align}
where $\tilde{\mathbf{x}_{i^{t}}}$ is the reconstruction of $\mathbf{x}_{i^{t}}$ produced by the autoencoder. This loss encourages the reconstruction to be similar to the sample's temporal neighbors, guiding the encoder to extract features that capture the temporally stationary component while ignoring irrelevant time-varying details.

{In our implementation, the temporal neighborhood $P_{i^{t}}$ is defined as the five samples centered around each timestamp $t$, specifically $\{t-2, t-1, t, t+1, t+2\}$. These temporal neighbors are treated as semantically similar to the center sample. This similarity definition is used solely to compute the temporal consistency loss and is independent of the actual batch construction used during training.}

\subsubsection{Feature Consistency}  
Feature consistency is inspired by the observation that different persons perform the same type of activity in different fashions, but different fashions do not hinder other people from identifying the activity type. Hence, we assume that the personal or individual features in the activity data may not be necessary in the activity clustering stage, and the features, which are commonly present across multiple data points, may be the essential features of the activity. The feature consistency loss function is based on this assumption. 

Previous research has demonstrated that combining carefully designed, handcrafted high-level features, which capture the essential characteristics of temporally varying signals, with the k-nearest neighbor algorithm can accurately classify sensor data associated with human activities \cite{pamap2}. Due to its effectiveness and simplicity, it is employed in this approach to define the local neighborhood of a data sample. The feature consistency loss then aims to preserve the high-level feature characteristics that are generally present in the local neighborhood. 

The feature consistency loss encourages the reconstruction of a data sample to be similar to its neighbors in the feature space. The rationale is that if data samples are close to each other in the handcrafted feature space, they likely represent the same type of activity. Thus, the features shared across multiple nearby data samples should represent the essential characteristics of that activity type. Features that only exist in some samples but not others likely represent individual or person-specific variations rather than core activity characteristics.

Formally, let $\mathbf{x}_{i^{f}}$ denote data sample $i$ in the feature space, and $\tilde{\mathbf{x}_{i^{f}}}$ the reconstruction of $\mathbf{x}_{i^{f}}$ produced by the autoencoder. Let $Q_{i^{f}}$ denote the index set of $|Q|$ local neighbors, $\mathbf{x}_{q}$, of $\mathbf{x}_{i^{f}}$ in the handcrafted feature space. The feature consistency loss $\Phi_{fc}$ for $\mathbf{x}_{i^{f}}$ is defined as:
\begin{align}
\Phi_{fc}(\mathbf{x}_{i^{f}}) = \frac{1}{|Q|}\sum_{q \in Q_{i^{f}}}||\mathbf{x}_{q} - \tilde{\mathbf{x}_{i^{f}}}||^{2}  \label{lossfc}
\end{align}

This loss encourages the reconstruction of the center point $\mathbf{x}_{i^{f}}$ to be similar to all of its neighbors $\mathbf{x}_{q}$ in the feature space. By minimizing the difference between the reconstruction of a sample and its neighbors, the model learns to capture the common features shared across similar activity examples while disregarding individual variations. This guides the encoder to focus on information that is consistent across the neighborhood rather than on unique characteristics of individual samples.

{For the feature consistency loss, we define the local neighborhood $Q_{i^{f}}$ using $k$-nearest neighbors in the handcrafted statistical feature space, where $k=5$. These feature neighbors are considered similar in terms of shared activity characteristics. As with temporal consistency, this neighbor selection is used exclusively for defining similarity relationships in the loss function and does not affect how training batches are constructed.}

\subsubsection{Model Architecture and Training Process}

\begin{figure}[H]

\includegraphics[width=0.7\textwidth]{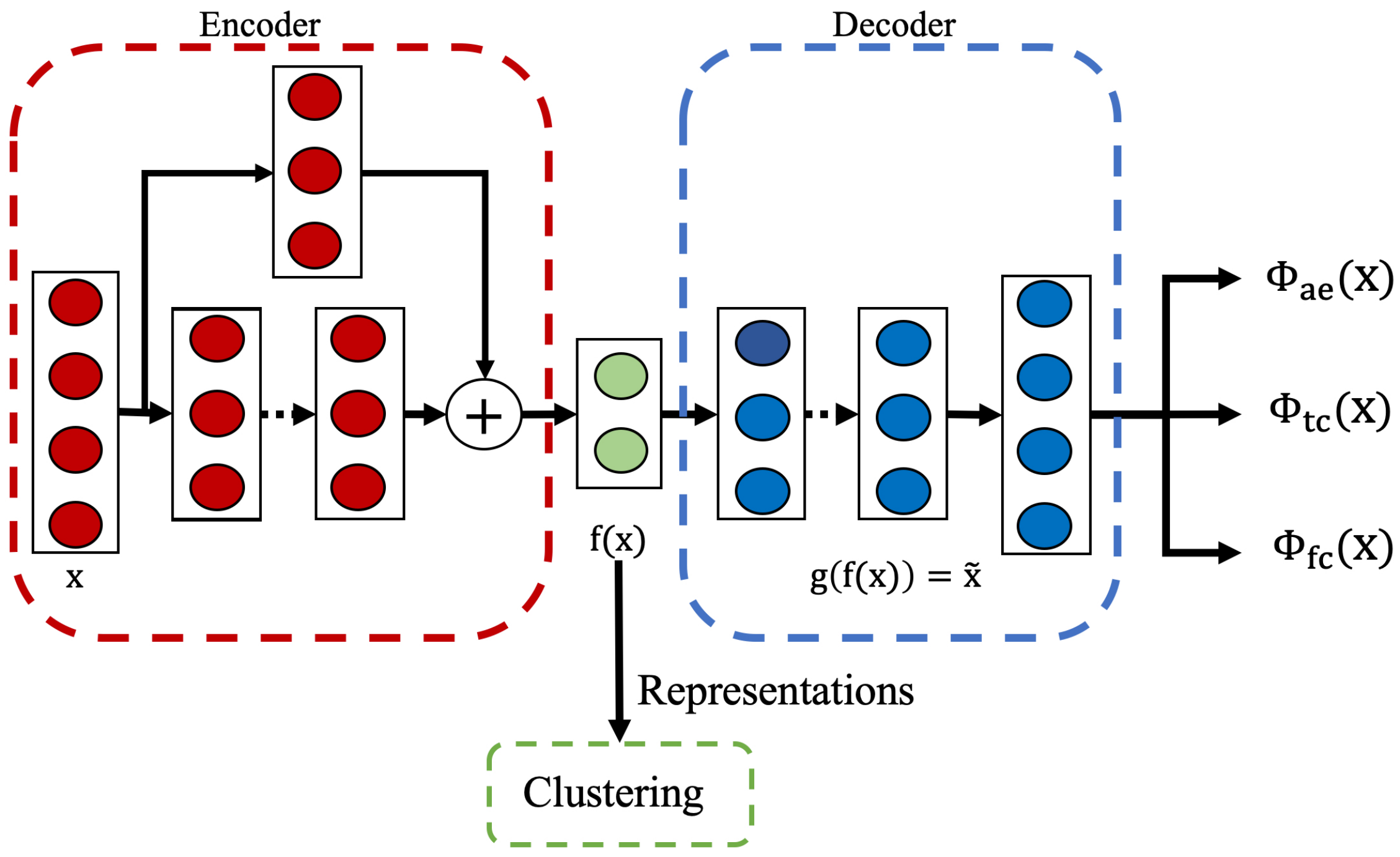} 
\caption{The residual autoencoder architecture used in the self-supervised model.} \label{resAE}

\end{figure}

As shown in Figure~\ref{resAE}, the foundation of the overall architecture of this approach is an autoencoder framework with two task-oriented objective functions $\Phi_{tc}(\mathbf{x}_{i})$, $\Phi_{fc}(\mathbf{x}_{i})$ and one regular reconstruction objective function $\Phi_{ae}(\mathbf{x}_i)$. By integrating guidance customized to HAR specifics, this approach facilitates learning representations that emphasize task-relevant information and minimize focus on irrelevant details. This equips the resulting representations with greater utility for subsequent activity analysis tasks.
\vspace{-6pt}

Specifically, the model is trained using a joint loss function combining temporal consistency, feature consistency, and reconstruction objectives. This composite loss is defined as Equation \eqref{joint1}:
\begin{align} \label{joint1}
\begin{split}
\Phi_{ss}(\mathbf{x}) = \sum_{i=0}^{N}(1-\alpha-\beta) \cdot \Phi_{ae}(\mathbf{x}_i) + \alpha \cdot \Phi_{tc}(\mathbf{x}_i) + \beta \cdot \Phi_{fc}(\mathbf{x}_i)
\end{split}
\end{align}
where $i$ is the index of the sample, $N$ is the size of the dataset, and $\alpha$ and $\beta$ are the parameters to balance the contribution of $\Phi_{ae}$, $\Phi_{tc}$, and $\Phi_{fc}$. While $\Phi_{tc}$ and $\Phi_{fc}$ preserve more task-relevant information in the representation, the $\Phi_{ae}$ component is also necessary in the learning process because without the reconstruction loss $\Phi_{ae}$, the risk of learning trivial solutions or worse representations will increase \cite{taxonomy}. By encouraging the model to accurately reconstruct the input data, $\Phi_{ae}$ compels the learned representations to maintain the underlying structure and properties inherent in the data during the training. The specific contributions of each loss component to the overall performance are further analyzed through ablation studies presented in Section~\ref{ablationsection}.

\subsection{Weakly Self-Supervised Approach}
Building on the strengths and limitations of the approaches presented so far, we now introduce a novel two-stage framework that integrates weakly supervised and self-supervised learning. This weakly self-supervised approach bridges the concepts presented in Sections \ref{weaksup} and \ref{ssa}, combining their complementary strengths to further reduce label dependency while maintaining strong performance.

Section \ref{weaksup} explores weak supervision using pairwise constraints with Siamese networks, while Section \ref{ssa} leverages self-supervision through domain knowledge losses to train a residual autoencoder without labeled data. The proposed approach integrates these methods by employing self-supervised training in the initial phase, followed by fine-tuning with limited pairwise constraints. Specifically, the first stage utilizes temporal and feature consistency losses from Section \ref{ssa} for self-supervised pre-training. In the second stage, the model is fine-tuned in a weakly supervised manner guided by similarity pairs, as outlined in Section \ref{weaksup}. By combining self-supervised residual autoencoder training with weakly supervised Siamese networks, this unified two-stage framework retains the advantages of self-supervision while effectively incorporating available weak supervision.

It is worth noting that we deliberately selected the weakly supervised single-task approach rather than the multi-task approach with self-supervision. This design choice stems from a fundamental compatibility consideration: the feature consistency component in our self-supervised approach explicitly attempts to eliminate person-specific variations by encouraging the model to focus on common activity patterns across different individuals. This objective would directly conflict with the person identification task in the multi-task approach, which aims to preserve and leverage precisely these individual differences. By using the single-task approach focused solely on activity recognition, we ensure alignment in our combined framework as both components share the goal of identifying activity patterns while disregarding person-specific variations.

\subsubsection*{Model Architecture and Training Process}
\begin{figure}[H]

\includegraphics[width=0.95\textwidth]{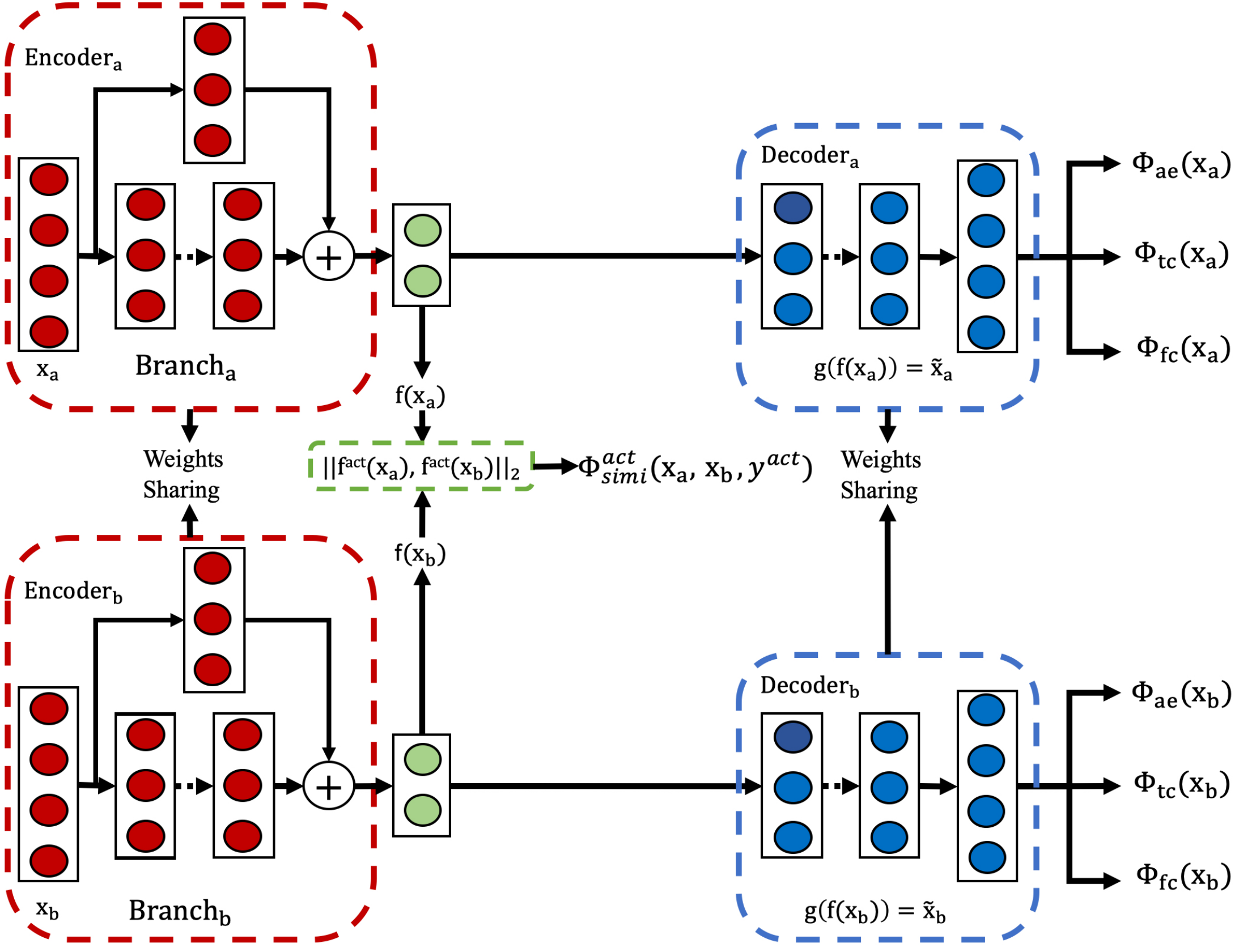} 
\caption{{The} 
 Siamese residual autoencoder architecture used in the weakly self-supervised model.} \label{resAEsiam}

\end{figure}

The architecture of the proposed weakly self-supervised model is depicted in Figure~\ref{resAEsiam}. The training process unfolds over two stages, each contributing uniquely to the model's learning capabilities.

In the first stage, the model is trained using the joint loss function defined in \mbox{Equation~\eqref{joint1}}. This loss function ensures that the model focuses on retaining task-relevant features while minimizing the influence of irrelevant ones. By leveraging self-supervised learning at this stage, the model establishes a foundational understanding of the underlying data structure.

In the second stage, the similarity loss $\Phi_{simi}^{act}$ is introduced to enhance the model's performance. Building upon the learned representations from the first stage, this phase applies a small set of pairwise constraints derived from weakly supervised data. These constraints refine the model's representations by aligning them more closely with the target classes. The joint loss function for this stage is formulated as Equation \eqref{joint2}:
\vspace{-6pt}
\begin{align} \label{joint2}
\begin{split}
&\Phi_{wss}(\mathbf{x}_{a}, \mathbf{x}_{b}, \mathbf{y}^{act}) = \sum_{i=0}^{N} (1-\alpha-\beta-\gamma) \cdot (\Phi_{ae}(\mathbf{x}_{a})+ \Phi_{ae}(\mathbf{x}_{b})) + \\
&\alpha \cdot (\Phi_{tc}(\mathbf{x}_{a}) + \Phi_{tc}(\mathbf{x}_{b})) + \\
&\beta \cdot (\Phi_{fc}(\mathbf{x}_{a}) + \Phi_{fc}(\mathbf{x}_{b})) + \\
&\gamma \cdot \Phi_{simi}^{act}(\mathbf{x}_{a}, \mathbf{x}_{b}, \mathbf{y}^{act})
\end{split}
\end{align}
where $\alpha$, $\beta$, and $\gamma$ are weight parameters that balance the contributions of temporal consistency, feature consistency, and similarity loss, respectively.

The similarity supervision $\Phi_{simi}^{act}$ plays a pivotal role in this stage. Despite being limited in quantity, it is highly relevant to the activity recognition task. To maximize its impact, the similarity loss is assigned a larger weight, ensuring it dominates the learning process. Conversely, the remaining loss functions (two self-supervised and one unsupervised) serve as regularization terms. Their inclusion prevents overfitting and promotes robust and generalized learning. Additionally, the model's feature extraction process remains consistent with the procedure outlined in Section \ref{featExtract}, maintaining methodological coherence across~stages.

\subsection{Methodology Summary and Clustering Approach}  
Table~\ref{techMethod} summarizes the technical methodologies employed by the discussed approaches. All models are trained via standard backpropagation with stochastic gradient descent, with parameters initialized using small random values. The supervised learning approach is optimized with cross-entropy loss using ground-truth labels, aiming primarily at accurate classification rather than the extraction of latent representations.

In contrast, the remaining approaches, including the unsupervised, self-supervised, and weakly supervised paradigms, as well as their extensions, are designed to learn mappings from input data $\mathbf{x}$ to latent representations $\mathbf{f}(\mathbf{x})$, such that semantically similar activities are embedded close to one another in the latent space. This clustering-friendly structure enables direct application of off-the-shelf clustering algorithms on the learned representations. In our experiments, we applied k-means clustering to the latent space to group similar activity samples into clusters.
\begin{table}[H]
\caption{Technical methodologies used in different approaches.}
\label{techMethod}

\begin{adjustwidth}{-\extralength}{0cm}
\begin{minipage}{\fulllength}
\begin{tabularx}{\textwidth}{LLLL}  
\toprule
\textbf{Approach} & \textbf{Model} &  \textbf{Loss} & \textbf{Input} \\

\midrule
{Supervised}
& {TCN + ResNet} 
& {$\Phi_{ce}(\mathbf{x}, \mathbf{y})$} 
& Raw data \\

\midrule

{Unsupervised}
& {Residual autoencoder} 
& {$\Phi_{ae}(\mathbf{x})$}
& Handcrafted features \\

\bottomrule
\end{tabularx}
\end{minipage}
\end{adjustwidth}

\end{table}

\begin{table}[H]\ContinuedFloat

\caption{{\em Cont.}}

\begin{adjustwidth}{-\extralength}{0cm}
\begin{minipage}{\fulllength}
\begin{tabularx}{\textwidth}{LLLL} 
\toprule
\textbf{Approach} & \textbf{Model} &  \textbf{Loss} & \textbf{Input} \\
\midrule

{Weakly supervised  single-task}
& {TCN + ResNet +  siamese networks} 
& {$\Phi_{simi}^{act}(\mathbf{x_{a}}, \mathbf{x_{b}}, \mathbf{y^{act}})$}
& Raw data \\

\midrule
{Weakly supervised  multi-task}
& {TCN + ResNet +  extended siamese networks} 
& {$\Phi_{mt-simi}(\mathbf{x_{a}}, \mathbf{x_{b}}, \mathbf{y^{act}}, \mathbf{y^{pers}})$} 
& Raw data \\

\midrule
{Self-supervised}
& {Extended  residual autoencoder} 
& {$\Phi_{ss}(\mathbf{x})$}
& Handcrafted features \\
\midrule
{Weakly self-supervised}
& {Extended  residual autoencoder +  siamese networks} 
& {$\Phi_{wss}(\mathbf{x_{a}}, \mathbf{x_{b}}, \mathbf{y^{act}})$} 
& Handcrafted features \\
\bottomrule
\end{tabularx}
\end{minipage}
\end{adjustwidth}
\end{table}

\section{Evaluation and Experiments}
We evaluated the effectiveness of our proposed approaches on three benchmark datasets: UCI-Smartphone \cite{sbhar}, PAMAP2 \cite{pamap2}, and REALDISP \cite{realdisp}, focusing on their performance in the HAR task with an emphasis on label efficiency. {Using both accuracy and F1-score as evaluation metrics, we compared our methods against established techniques as well as our own implemented baselines.}

For comparison with published works, we included several supervised approaches: DeepConvLSTM \cite{cnnLSTMHAR}, which integrates convolutional and LSTM recurrent units to capture temporal dynamics in sensor data without requiring manual feature engineering; Self-Attention \cite{DBLP:journals/corr/abs-2003-09018}, which employs attention mechanisms to learn the importance of different sensor modalities and better capture spatio-temporal context; ResGCNN \cite{Liao_2022}, which leverages residual graph convolutional neural networks to model relationships between sensor data and enable transfer learning capabilities; and Contrastive GNN \cite{tello2025contrasting}, which models spatial dependencies between sensors as a graph and applies contrastive learning to maximize mutual information between local and global representations. Additionally, we considered Efficient Deep Clustering \cite{pmlr-v189-mahon23a}, which offers a different approach by training an encoder through iterative pseudo-labeling, where temporally consistent clusters are refined using UMAP dimensionality reduction and HMM-based label smoothing. Most of these baseline methods, with the exception of Efficient Deep Clustering, rely on fully supervised learning paradigms, requiring complete labeled datasets for training, which highlights the importance of our research on reducing label dependency.

A consistent experimental protocol was followed across datasets, with stratified splits ensuring balanced activity class distributions in training, validation, and test sets. Beyond the metrics, we analyzed the quality of the learned representations, using visualization techniques to illustrate how effectively each method captured meaningful structures and discriminative features from raw sensor data. These comparisons highlight the advantages of the methods in producing task-relevant and label-efficient representations for HAR.

{To isolate the effect of supervision strategy from architectural complexity, we adopted consistent backbone encoders within each methodological family. Supervised and weakly supervised approaches share a TCN-based encoder, while the unsupervised and self-supervised models use a residual-based encoder, which is suited for processing handcrafted statistical features and incorporating domain-specific inductive biases. As described in Section}~\ref{weaksup}, {we also standardized the weak supervision signal using pairwise constraints across applicable methods. This design allows for a clean comparison of learning paradigms under aligned modeling conditions. The detailed model architectures and training configurations are provided in Appendix~\ref{appendix}.}

\subsection{Datasets} \label{datasets}

Our approaches were evaluated on benchmark HAR datasets featuring diverse activities performed by multiple individuals using various wearable sensors \cite{WANG20193}. These datasets introduce real-world challenges, including varied activities, sensing modalities, and intra- and inter-person variability. Sensors captured motion, orientation, and physiological signals from various body locations, including data from accelerometers, gyroscopes, magnetometers, and other devices.

To standardize input, raw sensor streams were pre-processed into fixed-size windows using dataset-specific sliding window techniques. This sensor-rich evaluation highlights the robustness, flexibility, and label efficiency of our methods in multimodal activity recognition under realistic conditions. Table~\ref{datasetSummary} summarizes the key characteristics of the datasets used in this study.

\begin{table}[H]
\centering
\caption{Details of the wearable-sensor-based activity recognition datasets.}
\label{datasetSummary}
\smallskip
\scalebox{0.9}{ 
\begin{tabular}{p{3cm}p{4.5cm}p{3.5cm}p{2.6cm}} 
\toprule
\textbf{\makecell{Dataset}} & \textbf{UCI-Smartphone} \cite{sbhar} & \textbf{PAMAP2} \cite{pamap2} & \textbf{REALDISP} \cite{realdisp} \\
\midrule
\makecell{Number of 
 \\ Volunteers} & 30 & 9 & 17 \\\midrule

\makecell{Sensors \\ Type and Number}
& \makecell[l]{Smartphone accelerometer: 1 \\ Smartphone gyroscope: 1} 
& \makecell[l]{IMUs: 3 \\ Heart rate monitor: 1} 
& IMUs: 9 \\
\midrule
Position of Sensors
& Waist 
& \makecell[l]{Dominant arm \\ Chest \\ Dominant side's ankle} 
& \makecell[l]{Back \\ L/R calf \\ L/R thigh \\ L/R lower arm \\ L/R upper arm} \\
\midrule
\makecell{Sampling \\ Frequency}
& 50 Hz 
& 100 Hz 
& 50 Hz \\
\midrule
\makecell{Activities \\ Number and Type}
& \makecell[l]{6 activities: \\ static daily living \\ dynamic daily living} 
& \makecell[l]{12 activities: \\ static daily living \\ dynamic daily living \\ household \\ sports} 
& \makecell[l]{33 activities: \\ fitness} \\
\bottomrule
\end{tabular}}
\end{table}

\subsubsection{UCI-Smartphone} 
The UCI-Smartphone dataset captures smartphone-based sensing data from 30 participants performing 12 activities. These include six fundamental motions (e.g., walking, lying down) and six postural transitions (e.g., stand-to-sit). Tri-axial accelerometer and gyroscope data were sampled at 50 Hz from smartphones worn on participants' waists, providing insights into both dynamic and static body behaviors. For this study, postural transitions were grouped into a combined transition class, and sensor streams were segmented into 2.56 s windows with a 1.28 s overlap, following established practices \cite{transitionAware}. This dataset presents challenges related to individual variations and mobile HAR scenarios.

\subsubsection{PAMAP2}
The PAMAP2 dataset includes multimodal sensor data from nine participants performing 12 diverse activities, ranging from sports exercises (e.g., rope jumping) to household tasks (e.g., vacuuming). Data were collected using a heart rate monitor and three IMUs positioned on the chest, dominant wrist, and ankle, measuring accelerometer, gyroscope, magnetometer, and temperature signals. Following prior work \cite{understandRNNHAR,dnnHARbenchmark}, sensor streams were downsampled from 100 Hz to 33.3 Hz and segmented into 5.12 s windows with a 1 s step size. This dataset provides comprehensive multimodal data for analyzing human activities across various contexts.

\subsubsection{REALDISP}
The REALDISP dataset provides sensor recordings from 17 participants performing 33 fitness and workout activities, {collected under different sensor placement conditions. In this study, we use the standard ideal-placement recordings to maintain consistency across all datasets. Other datasets in our evaluation, such as UCI-Smartphone and PAMAP2, do not include comparable displacement variants, so using the ideal setup ensures a clean and fair comparison of learning paradigms under aligned conditions.} The data were collected using 9 MTx inertial sensors placed on both arms, legs, and the torso, capturing 3D acceleration, angular velocity, and magnetic field orientation at a 50 Hz sampling rate. Sensor streams were segmented into non-overlapping 2 s windows, following standard practices. With its rich body-wide sensing and extensive activity diversity, REALDISP supports the development of robust techniques for HAR tasks \cite{realdisp}.

\subsection{Results and Analysis} \label{results}

The experimental results presented in Tables~\ref{weakSBHAR}--\ref{weakREALDISP} demonstrate the performance of the various deep learning approaches across the evaluated datasets. These findings offer a quantitative comparison, emphasizing the ability of the proposed methods to effectively capture meaningful activity representations. Notably, the proposed approaches achieve performance levels comparable to fully supervised models, despite relying on limited labeled data or weak supervision. In contrast, fully unsupervised models, which operate without any labeled data, exhibit significantly lower performance, underscoring the advantages of incorporating even minimal supervision for representation learning.

\begin{table}[H]
\caption{Results on UCI-Smartphone.} \label{weakSBHAR}
\scalebox{1}{%
\begin{tabular}{cccc}
\toprule
\textbf{Methods} &  \textbf{Acti. Acc.} &  \textbf{{Acti. F1.}} & \textbf{Pers. Acc} *\\
\midrule
        DeepConvLSTM \cite{cnnLSTMHAR} & 0.8942 &  {0.8782} & -\\
        Self-Attention \cite{DBLP:journals/corr/abs-2003-09018} & 0.8500 &  {0.8500} & -\\
        ResGCNN \cite{Liao_2022} & 0.8313 & {0.8333} & -\\
        Contrastive GNN \cite{tello2025contrasting} & 0.9143 & {0.9182} & -\\
        Efficient Deep Clustering \cite{pmlr-v189-mahon23a} & 0.6590 & {0.6440} & -\\
	Fully Supervised TCN & 0.9866 & {0.9832} & - \\
	Autoencoder  & 0.6369 & {0.6230} & -  \\
	Weakly Supervised Single-Task Approach  & 0.9436 & {0.9389} & - \\
	Weakly Supervised Multi-Task Approach   & 0.9885 & {0.9839} & 0.8892 \\
	Self-Supervised Approach  & 0.7401 & {0.7285} & - \\
	Weakly Self-Supervised Approach with 1\% Labels  & 0.8015 & {0.7926} & - \\
	Weakly Self-Supervised Approach with 5\% Labels  & 0.8853 & {0.8776} & - \\
	Weakly Self-Supervised Approach with 10\% Labels  & 0.9352 & {0.9290} & - \\
\bottomrule
\end{tabular}}
\noindent{\footnotesize{* Person identification metrics are only reported for the weakly supervised multi-task approach, as other methods were not designed for person identification.}}
\end{table}

{The performance of each method is influenced by dataset-specific characteristics that also mirror challenges encountered in real-world deployment. PAMAP2 consistently yields high accuracy for label-efficient approaches, including unsupervised, self-supervised, weakly supervised single-task, and multi-task methods, likely due to its favorable balance of sensor richness, activity diversity, and manageable complexity. REALDISP, while offering the most comprehensive body-wide sensing setup, includes a larger number of activity classes and higher inter-subject variability, which can increase learning difficulty, particularly under weak or self-supervised conditions. UCI-Smartphone, with its simpler sensor configuration and fewer activity classes, performs best under full supervision, where clean labels compensate for limited sensor coverage. These patterns highlight that sensor richness and dataset complexity, while potentially beneficial, may also introduce confounding factors such as user diversity, placement inconsistency, and signal noise. To ensure controlled comparisons across methods, this study focuses on ideal-placement settings, while recognizing that real-world deployments introduce additional variability not captured in this setup.}
\begin{table}[H]
\caption{Results on PAMAP2.} \label{weakPAMAP2}
\begin{tabularx}{\textwidth}{cCCC}
\toprule
\textbf{Methods} & \textbf{Acti. Acc.} &  \textbf{{Acti. F1.}} &  \textbf{Pers. Acc} *\\
\midrule
        DeepConvLSTM \cite{cnnLSTMHAR} & 0.8053 & {0.7792}  & -\\
        Self-Attention \cite{DBLP:journals/corr/abs-2003-09018} & 0.8300 &  {0.8100}  & -\\
        ResGCNN \cite{Liao_2022} & 0.8218 &  {0.8197}  & - \\
        Contrastive GNN \cite{tello2025contrasting} & 0.8686 & {0.8636}  & -\\

          Efficient Deep Clustering \cite{pmlr-v189-mahon23a} & 0.8630 & {0.8010}  & -\\
	Fully Supervised TCN & 0.9527 & {0.9382}  & - \\

         	Autoencoder  & 0.7706 & {0.7398}  & -  \\

	Weakly Supervised Single-Task Approach   & 0.9856 & {0.9736}  & - \\

        	Weakly Supervised Multi-Task Approach   & 0.9893 & {0.9766}  & 0.9881 \\

	Self-Supervised Approach & 0.8543 & {0.8259}  & - \\
	Weakly Self-Supervised Approach with 1\% Labels & 0.9128 & {0.8967}  & - \\
	Weakly Self-Supervised Approach with 5\% Labels & 0.9755 & {0.9631}  & - \\
	Weakly Self-Supervised Approach with 10\% Labels & 0.9904 & {0.9782}  & - \\
\bottomrule

\end{tabularx}
\noindent{\footnotesize{* Person identification metrics are only reported for the weakly supervised multi-task approach, as other methods were not designed for person identification.}}
\end{table}
\unskip
\begin{table}[H]
\caption{Results on REALDISP.} \label{weakREALDISP}
\scalebox{1}{%
\begin{tabular}{cccc}
\toprule
\textbf{Methods} & \textbf{Acti. Acc.} &  \textbf{{Acti. F1.}} &  \textbf{Pers. Acc} *\\
\midrule
        DeepConvLSTM \cite{cnnLSTMHAR} & 0.8826 &  {0.8934} & - \\
        Self-Attention \cite{DBLP:journals/corr/abs-2003-09018} & 0.7200 & {0.6800}  & - \\
        ResGCNN \cite{Liao_2022} & 0.7775 &  {0.7435}  & - \\
        Contrastive GNN \cite{tello2025contrasting} & 0.9365 &  {0.9152}  & - \\
        Efficient Deep Clustering \cite{pmlr-v189-mahon23a} & 0.9100 & {0.8800}  & -\\
	Fully Supervised TCN & 0.9710 & {0.9511}  & - \\
	Autoencoder  & 0.6401 & {0.6243} & - \\
	Weakly Supervised Single-Task Approach   & 0.9796 & {0.9583}  & - \\
	Weakly Supervised Multi-Task Approach   & 0.9822 & {0.9616}  & 0.9668 \\
	Self-Supervised Approach & 0.6812 & {0.6569}  & - \\
	Weakly Self-Supervised Approach with 1\% Labels & 0.7515 & {0.7351}  & - \\
	Weakly Self-Supervised Approach with 5\% Labels & 0.8780 & {0.8578}  & - \\
	Weakly Self-Supervised Approach with 10\% Labels & 0.9223 & {0.9027}  & - \\
\bottomrule
\end{tabular}}
\noindent{\footnotesize{* Person identification metrics are only reported for the weakly supervised multi-task approach, as other methods were not designed for person identification.}}
\end{table}


One of the key advantages of non-fully supervised learning approaches is their ability to significantly reduce the reliance on explicitly labeled data. However, this label efficiency comes with an inherent trade-off in accuracy. The results indicate that fully supervised models generally achieve consistently strong performance because of their direct access to complete labeled ground truth. In contrast, weakly supervised and self-supervised methods must infer structure and relationships from limited annotations or proxy tasks, which can introduce inconsistencies and errors. This trade-off becomes particularly evident in the weakly self-supervised approach, where increasing the fraction of labeled data from 1\% to 10\% leads to improved accuracy, though still falling short of fully supervised methods. This demonstrates that achieving label efficiency requires a careful balance between minimizing supervision requirements and maintaining model reliability.

The following subsections provide a detailed analysis of these outcomes, examining how each model performs and the insights derived from their results.

\subsubsection{Weakly Supervised Single-Task Approach}
The weakly supervised single-task approach demonstrates strong performance across all datasets, achieving 94.36\%, 98.56\%, and 97.96\% accuracy on UCI-Smartphone, PAMAP2, and REALDISP, respectively. While it does not consistently outperform all supervised methods, it achieves results comparable to or better than several fully supervised techniques despite not requiring explicitly labeled data during training. By incorporating limited supervision in the form of activity similarity information, this method significantly outperforms unsupervised approaches (which achieve only 63.69\%, 77.06\%, and 64.01\% across the datasets). These findings underline the potential of weak supervision to bridge the gap between fully supervised and unsupervised learning paradigms.

\subsubsection{Weakly Supervised Multi-Task Approach}
The weakly supervised multi-task approach achieves exceptional performance across all three datasets (98.85\%, 98.93\%, and 98.22\%), outperforming our weakly supervised single-task method and many fully supervised approaches, including DeepConvLSTM, Self-Attention, and ResGCNN. On the UCI-Smartphone dataset, it even exceeds the performance of our fully supervised TCN baseline (98.85\% vs. 98.66\%). By leveraging limited supervision that accounts for both activity and person similarity, this method enhances performance by disentangling semantic representations while maintaining similarity metrics within the activity and person domains. Its superior performance compared to single-task approaches highlights the framework's effective capability to share information among related tasks, illustrating the advantages of multi-task learning in weakly supervised settings.

\subsubsection{Self-Supervised Approach}
The self-supervised approach, an enhanced version of the standard autoencoder, consistently outperforms the standard autoencoder across all three datasets (74.01\% vs. 63.69\% on UCI-Smartphone, 85.43\% vs. 77.06\% on PAMAP2, and 68.12\% vs. 64.01\% on REALDISP). It effectively derives meaningful features for activity clustering without relying on labeled data. Although fully supervised methods achieve higher accuracy by leveraging labeled data, they are constrained by the need for such annotations. The performance gap between self-supervised and fully supervised approaches varies by dataset, with PAMAP2 showing the smallest gap (85.43\% vs. 95.27\%). {We note that activity transitions, such as sit-to-stand or stand-to-walk, tend to produce ambiguous signal patterns that may challenge the assumption of temporal continuity. These edge cases can introduce inconsistencies in representation learning, limiting the effectiveness of self-supervised objectives that rely on local consistencies. Despite these limitations, the approach successfully learns robust representations that enhance clustering performance, showcasing its potential in label-scarce scenarios.}


\subsubsection{Weakly Self-Supervised Approach}
The weakly self-supervised approach achieves significant improvements over both unsupervised and self-supervised methods, underscoring the benefits of integrating even minimal weakly labeled data. Its performance demonstrates a clear trend: as the amount of labeled data increases, so does the model's effectiveness. With just 1\% of labeled data, the approach already achieves substantial gains over the purely self-supervised method (80.15\% vs. 74.01\% on UCI-Smartphone, 91.28\% vs. 85.43\% on PAMAP2, and 75.15\% vs. 68.12\% on REALDISP). With 10\% of labeled data, it approaches the performance of many fully supervised methods (93.52\%, 99.04\%, and 92.23\% across the three datasets), despite using only a small fraction of the annotations. In particular, on PAMAP2, with 10\% labels it achieves 99.04\% accuracy, exceeding even our fully supervised TCN baseline (95.27\%). These results validate our design decision to integrate the single-task approach with self-supervision rather than the multi-task approach, as the alignment between their objectives (both focusing on activity recognition while disregarding person-specific variations) enables effective knowledge transfer even with minimal labeled data. This ability to maximize the utility of available information underscores its adaptability and efficiency. Overall, the results validate the strength of this approach in effectively combining weak supervision with self-supervision to outperform purely unsupervised and self-supervised alternatives.

\subsection{Visualization} \label{vis} 
\begin{figure}[H]
  
    \begin{minipage}{0.3\textwidth}
        \centering
        \includegraphics[width=\textwidth]{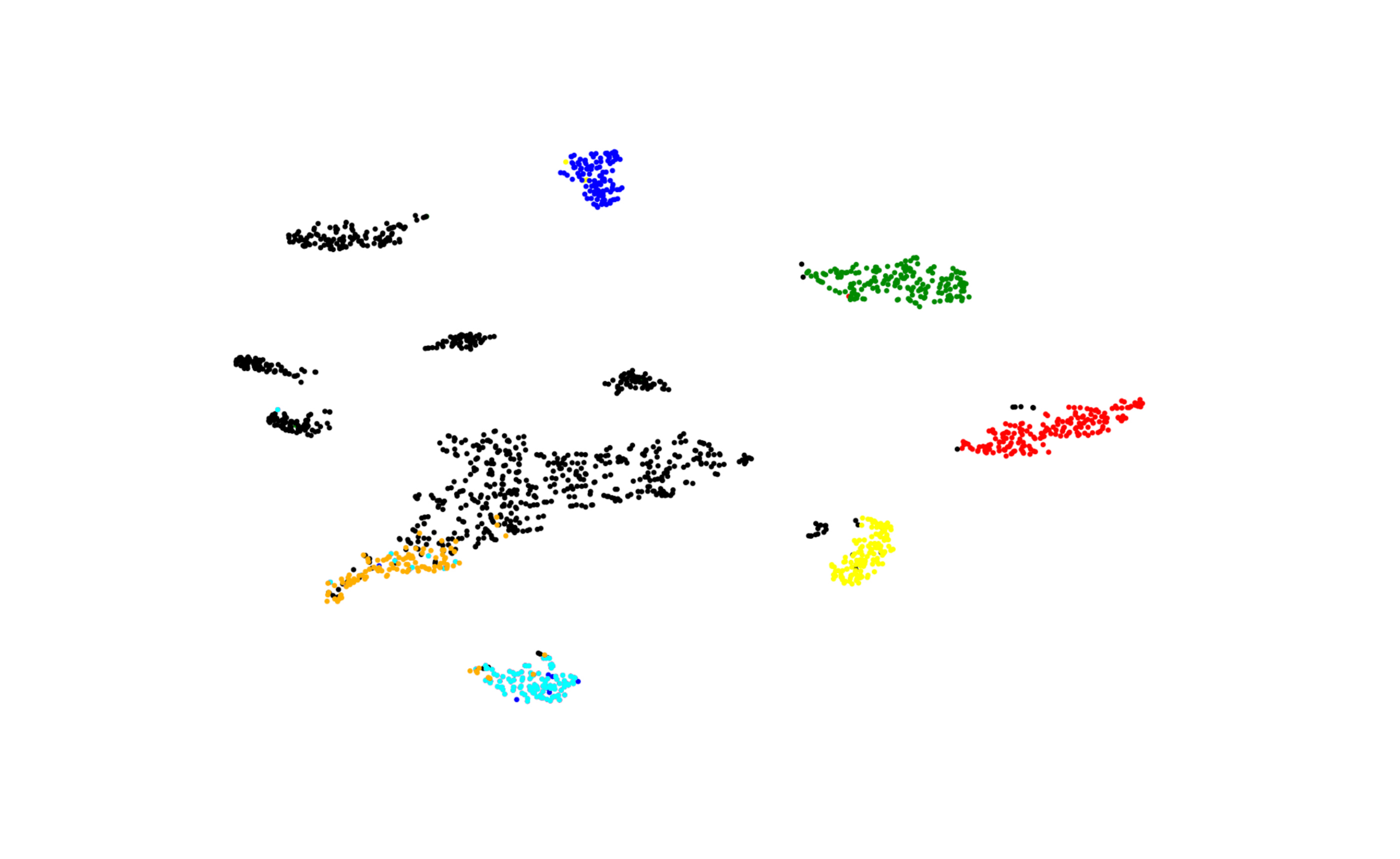}
        \caption*{\centering (\textbf{a})}
        \label{SBHARwss}
    \end{minipage}
    \hspace{0.1em} 
    \begin{minipage}{0.3\textwidth}
        \centering
        \includegraphics[width=\textwidth]{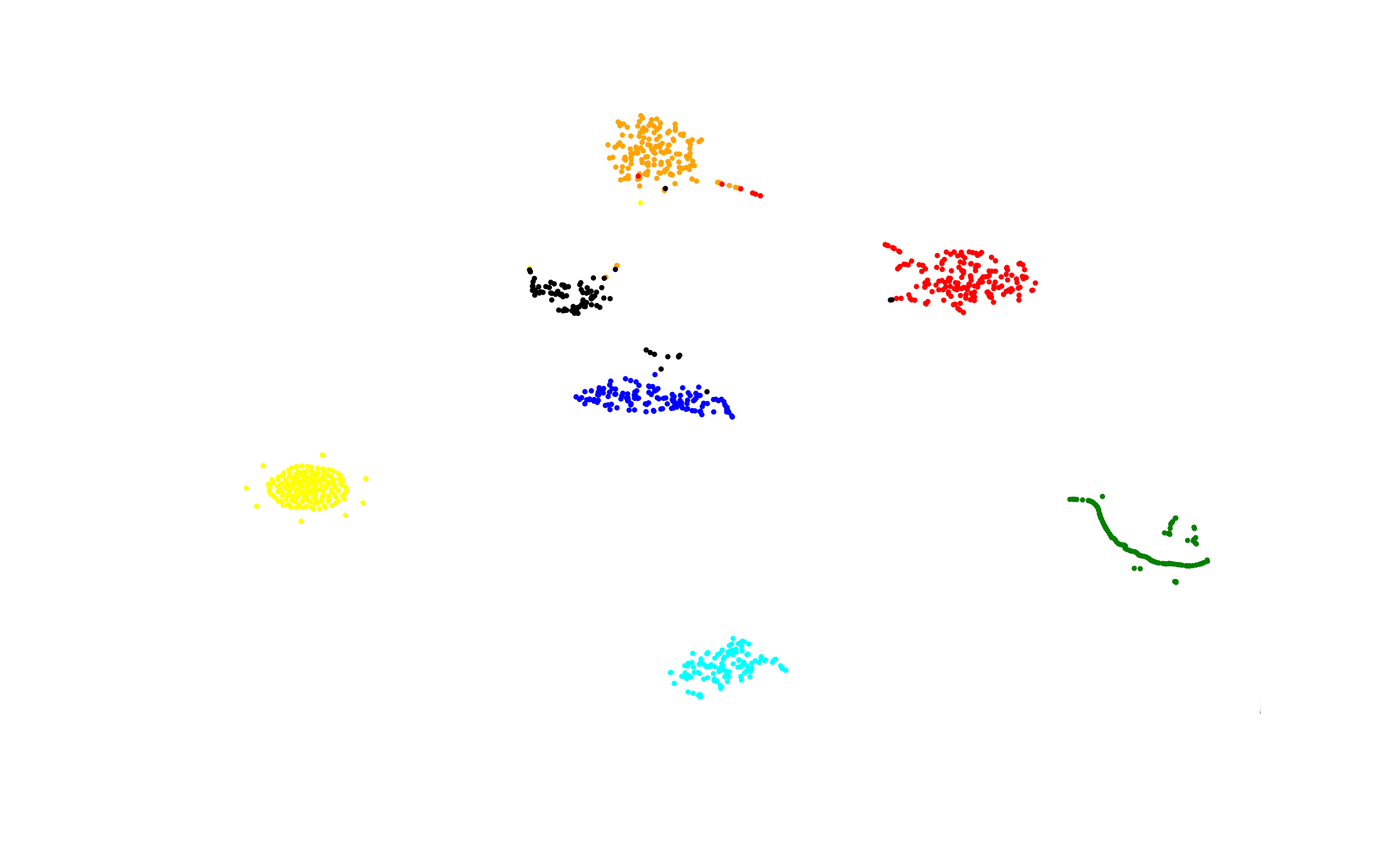}
        \caption*{\centering (\textbf{b})}
        \label{SBHARwsm}
    \end{minipage}
    \hspace{0.1em} 
    \begin{minipage}{0.3\textwidth}
        \centering
        \includegraphics[width=\textwidth]{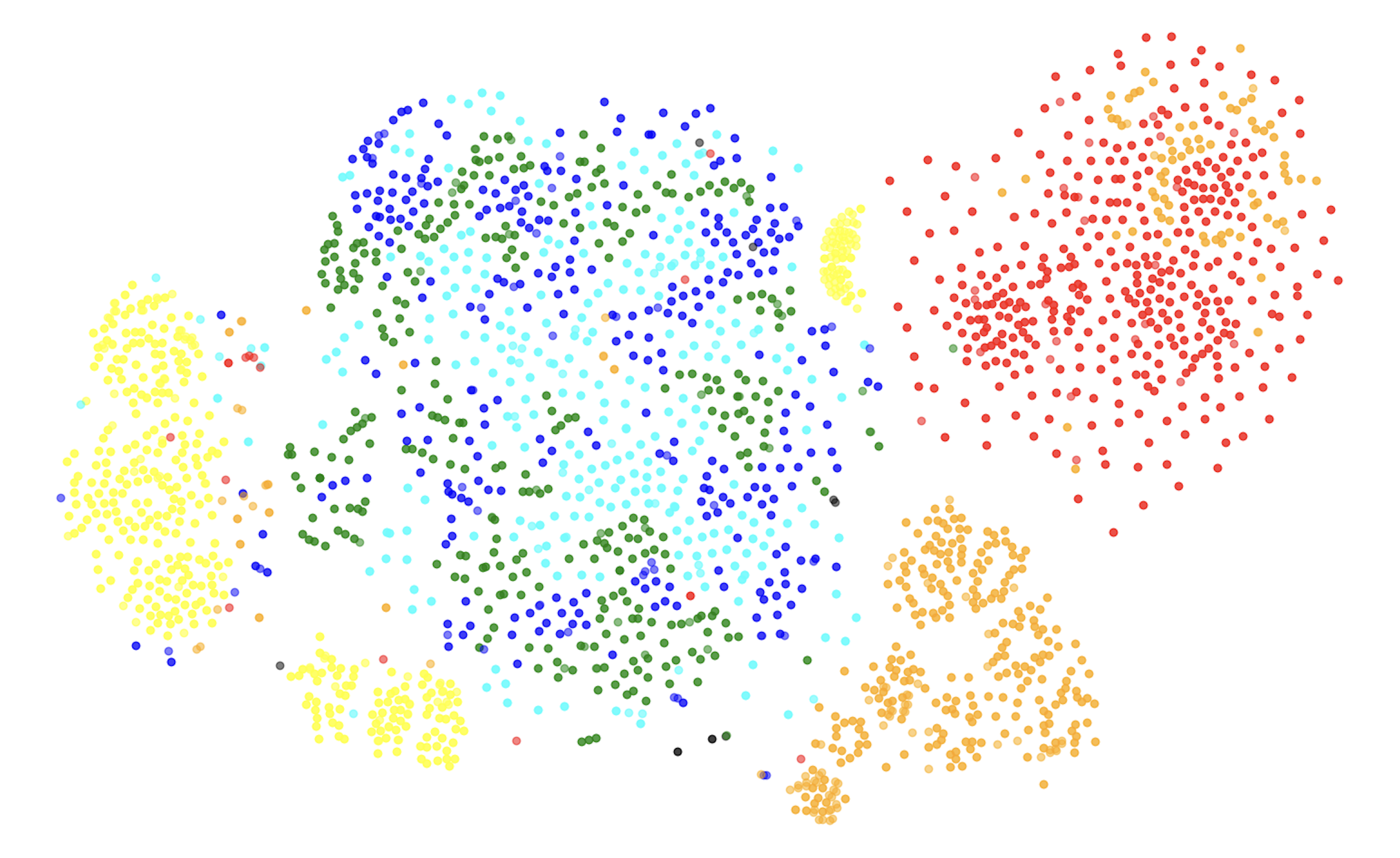}
        \caption*{\centering (\textbf{c})}
        \label{SBHARss}
    \end{minipage}
    \vspace*{0.1in}
    \caption{Visualization 
 of representation vectors learned by weakly supervised single-task, weakly supervised multi-task, and self-supervised approaches on the UCI-Smartphone dataset. Different colors indicate different activity types in the dataset. (\textbf{a}) Weakly 
 supervised single-task approach. (\textbf{b}) Weakly supervised multi-task approach. (\textbf{c}) Self-supervised approach.}
    \label{sbhar1}
\end{figure}

\begin{figure}[H]

    \begin{minipage}{0.3\textwidth}
        \centering
        \includegraphics[width=\textwidth]{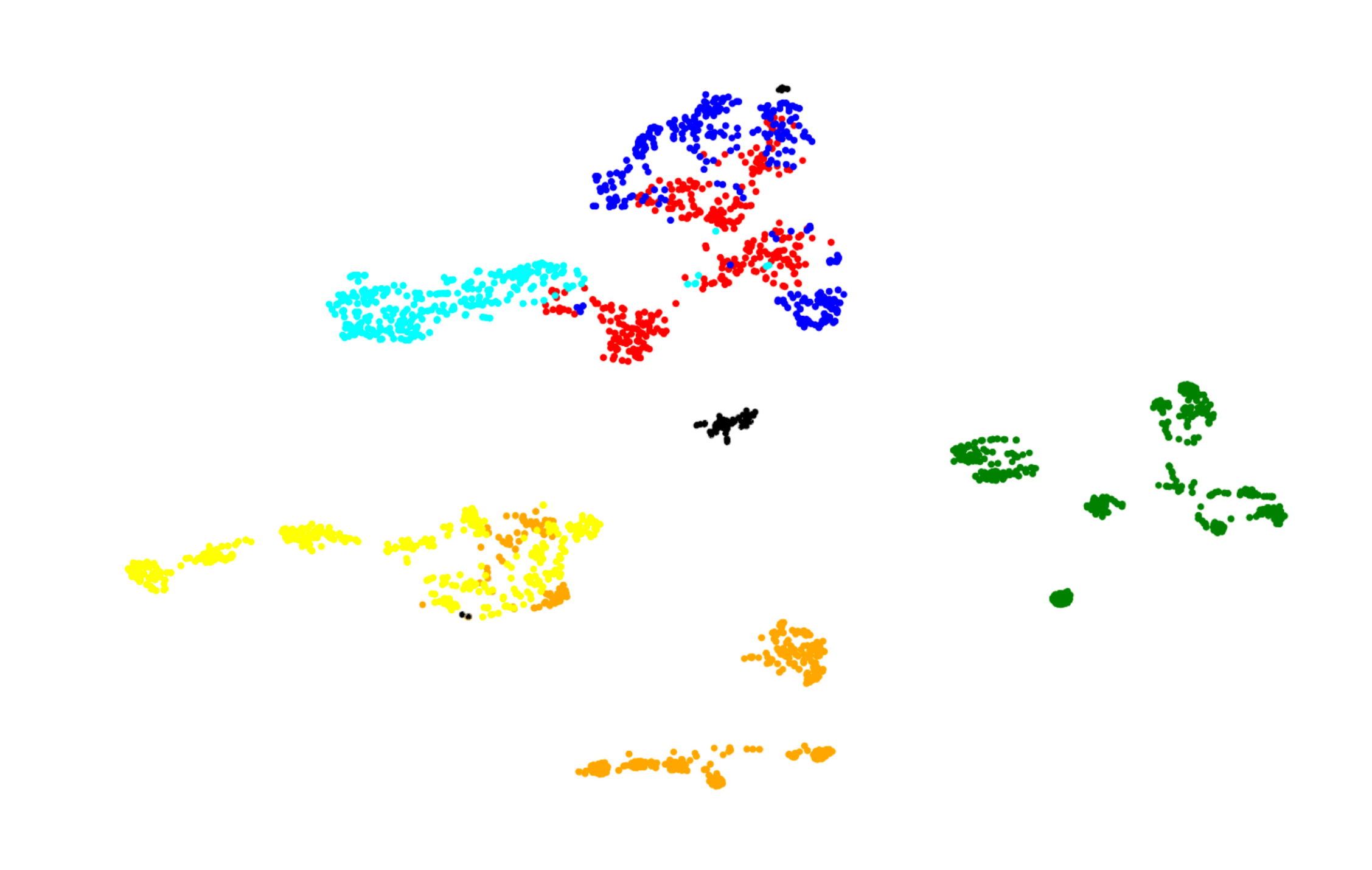}
        \caption*{\centering (\textbf{a})}
        \label{SBHAR1percent}
    \end{minipage}
    \hspace{0.1em} 
    \begin{minipage}{0.3\textwidth}
        \centering
        \includegraphics[width=\textwidth]{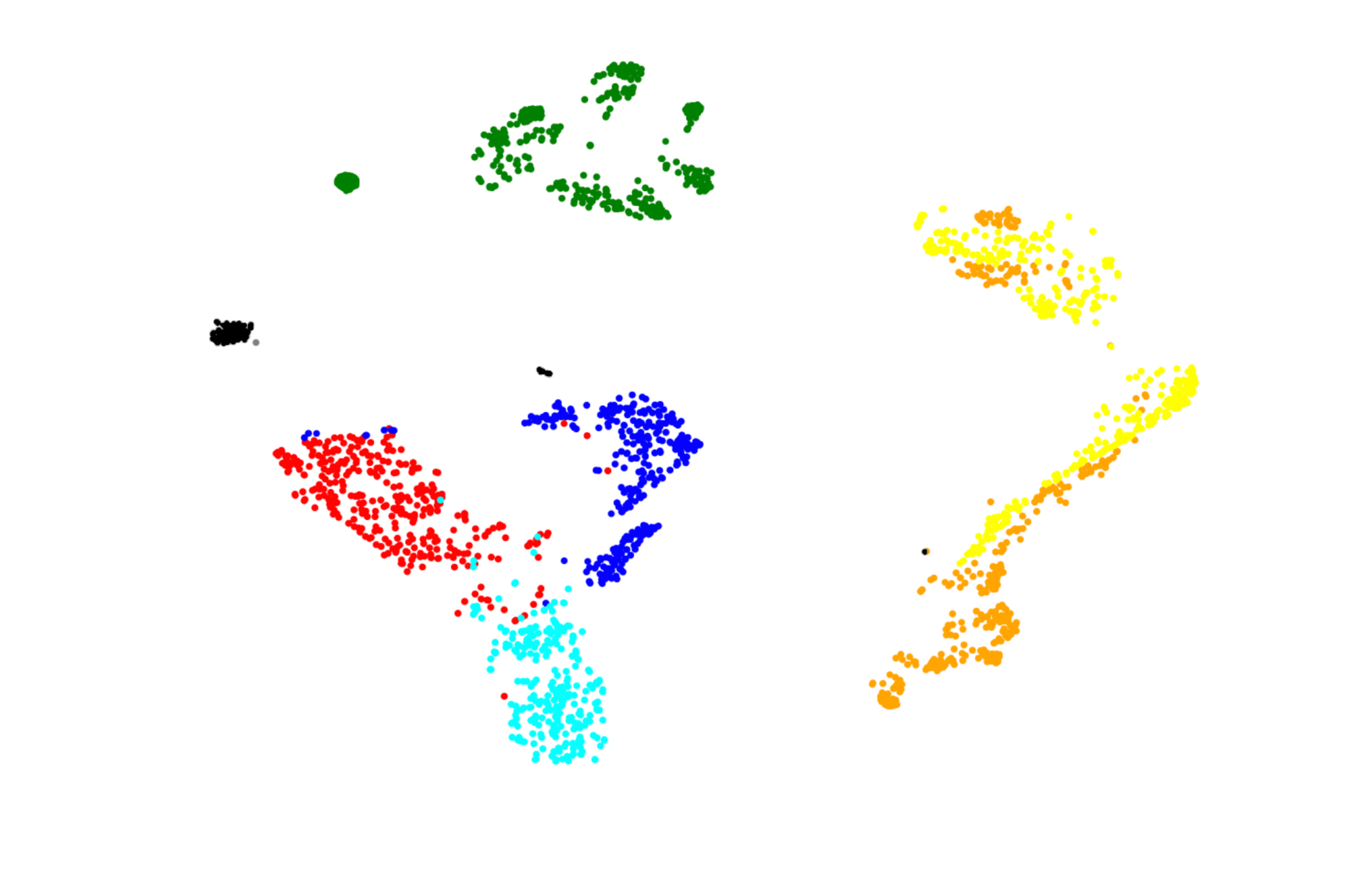}
        \caption*{\centering (\textbf{b})}
        \label{SBHAR5percent}
    \end{minipage}
    \hspace{0.1em} 
    \begin{minipage}{0.3\textwidth}
        \centering
        \includegraphics[width=\textwidth]{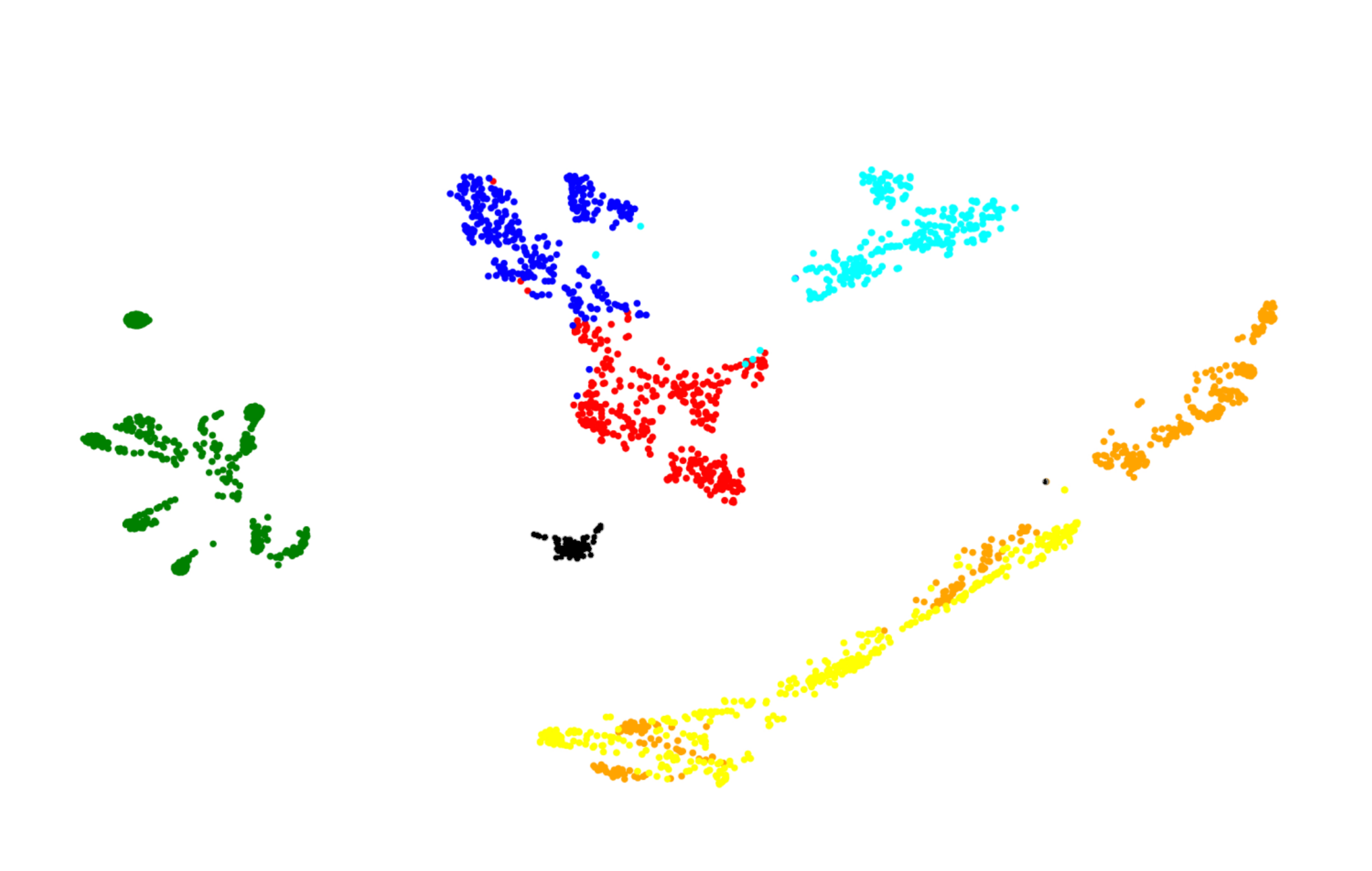}
        \caption*{\centering (\textbf{c})}
        \label{SBHAR10percent}
    \end{minipage}
    \vspace*{0.1in}
    \caption{Visualization 
 of representation vectors learned by the weakly self-supervised approach on the UCI-Smartphone dataset, using 1\%, 5\%, and 10\% of labeled data during training. Different colors indicate different activity types in the dataset. (\textbf{a}) Weakly self-supervised approach with 1\% labeled data.  (\textbf{b}) Weakly self-supervised approach with 5\% labeled data. (\textbf{c}) Weakly self-supervised approach with 10\% labeled data.}
    \label{sbhar3}
\end{figure}

To better understand the learned representations, we visualized the distribution of activity representation vectors using t-SNE \cite{tsne}, a technique that maps high-dimensional data into a low-dimensional space. These visualizations, derived from the UCI-Smartphone dataset, are presented in Figures~\ref{sbhar1} and \ref{sbhar3}, where different colors denote various activities. The following observations can be drawn:

\begin{enumerate}
\item Semantic disentanglement: The evaluated deep representation learning approaches effectively disentangle semantic representations, forming distinct clusters in the representation space. This separation facilitates downstream tasks such as clustering, which leverage these representations for improved performance.

\item Effectiveness of weakly supervised approaches: As shown in Figure~\ref{sbhar1}a,b, the weakly supervised single-task and multi-task approaches produce the most distinct clusters, benefiting from the use of all labeled data. However, this comes with the trade-off of requiring extensive labeling.

\item Multi-task mechanism advantage: The weakly supervised multi-task approach (\mbox{Figure~\ref{sbhar1}b}) outperforms the single-task variant (Figure~\ref{sbhar1}a) by leveraging knowledge sharing between activity recognition and person identification tasks, resulting in improved cluster quality.

\item Self-supervised efficiency: Figure~\ref{sbhar1}c illustrates the self-supervised approach’s effectiveness in generating reasonably distinct clusters. This highlights the role of temporal and feature consistency in shaping meaningful clusters.

\item Reduced labeling requirements: Although the weakly self-supervised approach (\mbox{Figure~\ref{sbhar3}}) does not achieve the same level of cluster distinctiveness as weakly supervised methods, it relies on only a fraction of the labeled data, whereas weakly supervised methods utilize the entire labeled dataset. This distinction underscores the weakly self-supervised approach's ability to alleviate the labeling burden while still maintaining competitive performance.
\end{enumerate}
\vspace{-12pt}

\subsection{Ablation Studies on the Effect of Domain Knowledge} \label{ablationsection}
To evaluate the individual contributions of different loss terms in our self-supervised approach, we conducted a series of ablation experiments by selectively removing components of the loss formulation. We focused on the self-supervised model as it incorporates multiple domain-specific objectives that can be independently analyzed.

Specifically, we trained separate models using reduced loss configurations: (i) a model trained with only the temporal consistency loss ($\Phi_{tc}$) and reconstruction losses ($\Phi_{ae}$), and (ii) a model trained with only the feature consistency loss ($\Phi_{fc}$) and reconstruction losses. By comparing the performance of these ablated models, we isolated the impacts of $ \Phi_{tc}$ and $ \Phi_{fc}$. The results of these experiments are presented in Table~\ref{ablation}.

\begin{table}[H]
    \caption{Ablation studies on the effect of each loss term in the self-supervised approach.}\label{ablation}
   
    \begin{tabularx}{\textwidth}{CC}
      \midrule
      \textbf{Loss Included} & \textbf{ACC}  \\
     \midrule
           \multicolumn{2}{c}{UCI-Smartphone 
} \\
      \midrule
      $ \Phi_{ae} $ & 0.6369   \\
      $ \Phi_{tc} + \Phi_{ae} $ & 0.6449   \\
      $ \Phi_{fc} + \Phi_{ae} $ & 0.7308    \\
      $ \Phi_{tc} + \Phi_{fc} + \Phi_{ae} $ & 0.7401  \\
      \midrule
      \multicolumn{2}{c}{PAMAP2} \\
      \midrule
      $ \Phi_{ae} $ & 0.7706   \\
      $ \Phi_{tc} + \Phi_{ae} $ & 0.7859   \\
      $ \Phi_{fc} + \Phi_{ae} $ & 0.8065   \\
      $ \Phi_{tc} + \Phi_{fc} + \Phi_{ae} $ & 0.8543  \\
      \midrule
      \multicolumn{2}{c}{REALDISP} \\
      \midrule
      $ \Phi_{ae} $ & 0.6401   \\
      $ \Phi_{tc} + \Phi_{ae} $ & 0.6341    \\
      $ \Phi_{fc} + \Phi_{ae} $ & 0.6610    \\
      $ \Phi_{tc} + \Phi_{fc} + \Phi_{ae} $ & 0.6812   \\
      \midrule
    \end{tabularx}
\end{table}
The findings reveal that models trained with either $\Phi_{fc} + \Phi_{ae}$ or $\Phi_{tc} + \Phi_{ae}$ outperform the vanilla autoencoder in most scenarios. However, the model incorporating all three loss terms, $\Phi_{tc} + \Phi_{fc} + \Phi_{ae}$, achieves the best overall results, indicating the complementary nature of $\Phi_{tc}$ and $\Phi_{fc}$. A closer examination highlights the greater impact of the $\Phi_{fc}$ loss, which aligns with the intuition that consistency in the feature space provides a stronger supervisory signal for activity recognition than temporal consistency alone.

Interestingly, we observed that on the REALDISP dataset, adding only temporal consistency ($\Phi_{tc}$) to the reconstruction loss ($\Phi_{ae}$) resulted in a slight performance decrease (from 64.01\% to 63.41\%). This dataset-specific behavior might be attributed to the unique characteristics of the REALDISP dataset, which contains 33 fitness activities with potentially more complex temporal dynamics than the other datasets. Nevertheless, the complete model with all three losses still achieves the best performance across all datasets, confirming the overall benefit of our approach.

These ablation studies underscore the effectiveness of incorporating domain knowledge into representation learning models. By dissecting the role of each loss component, we demonstrate that integrating these design choices enables models to achieve strong accuracy while significantly reducing the reliance on labeled data.

\section{Conclusions}
This study provides a comprehensive exploration across the supervision spectrum for HAR using wearable sensors, directly addressing the critical challenge of label dependency. We systematically investigated fully supervised learning, unsupervised learning, weakly supervised learning with pairwise constraints, multi-task learning with knowledge sharing, self-supervised learning incorporating domain knowledge, and our novel weakly self-supervised framework that combines complementary strengths of multiple paradigms. Through systematic evaluation across multiple benchmark datasets, we demonstrated how each approach navigates the fundamental trade-off between supervision requirements and recognition performance.

Our findings reveal several significant insights:  
First, weakly supervised methods achieve performance comparable to fully supervised approaches while eliminating the need for explicit activity labels, instead relying on pairwise similarity constraints that are often easier to obtain.  
Second, multi-task learning effectively integrates related tasks such as HAR and person identification, leveraging shared knowledge to enhance overall performance beyond what single-task approaches can achieve.  
Third, incorporating domain knowledge through temporal and feature consistency enables self-supervised approaches to learn meaningful representations without any labels, significantly outperforming standard unsupervised methods.  
Finally, our novel weakly self-supervised framework demonstrates remarkable label efficiency, achieving 99.04\% accuracy on PAMAP2 with only 10\% of labeled data, surpassing even fully supervised techniques that require complete label sets.

Our careful design choices, such as integrating single-task rather than multi-task supervision with self-supervision to ensure objective alignment, further enhanced the effectiveness of our approaches. These results have important implications for both research and practical HAR applications where labeled data is scarce or expensive to obtain. They provide a systematic framework for selecting appropriate learning methodologies based on label availability and application requirements. By reducing the annotation burden while maintaining high accuracy, this work advances representation learning for wearable sensor data and paves the way for more scalable, efficient, and practical HAR systems in real-world settings. Future work will consider expanding evaluation to additional datasets and deployment scenarios to further assess generalizability under varied sensor conditions.



 \vspace{6pt} 





\authorcontributions{Conceptualization, T.S. and M.H.; methodology, T.S. M.H.; software, T.S.; validation, T.S., M.H.; Writing---original draft preparation, T.S.; writing---review and editing, T.S. M.H.; visualization, T.S.; supervision, M.H. All authors have read and agreed to the published version of the manuscript.}

 \funding{This research received no external funding}

\conflictsofinterest{The authors declare no conflicts of interest.}





\appendixtitles{yes} 
\appendixstart

\appendix
\section{Model Architectures and Training Configurations}\label{appendix}
{This 
 appendix summarizes the network architectures and training settings used across the learning paradigms evaluated.}

\subsection{Backbone Architectures}
\begin{itemize}
  \item Residual 
 Autoencoder (unsupervised, self-supervised, weakly self-supervised): {Encoder: four-layer MLP with hidden sizes [256, 256, 128, 96], batch normalization, and ReLU. A skip connection adds the linearly projected input to the output. Decoder mirrors the encoder with layers [96, 128, 256, 256], ending in feature reconstruction.}
  
  \item Temporal Convolutional Network (TCN, supervised and weakly supervised): {Encoder: three stacked TCN blocks, each with three dilated 1D convolutions (kernel size 3, filters 128), followed by batch normalization and dropout (rate = 0.1). Skip connections within blocks. Global pooling outputs a 96-dim embedding.}
  
  \item Siamese Network (weakly supervised and weakly self-supervised): {Two weight-sharing branches reuse the TCN encoder, each followed by a fully connected layer producing a 96-dim representation. Contrastive loss is applied to the distance between~outputs.}
\end{itemize}

\subsection{Learning-Paradigm-Specific Configurations}
\begin{itemize}
  \item Supervised: {Cross-entropy loss over activity labels; Adam optimizer with learning rate $10^{-3}$; batch size 128.}
  
  \item Unsupervised: {Residual autoencoder trained with mean squared error on handcrafted features.}
  
  \item Weakly Supervised (Single/Multi-Task): {Contrastive loss with margin $\delta=1.0$. Multi-task: separate FC layers for activity and person embeddings; loss weights $\alpha=0.6$, $\beta=0.4$.}
  
  \item Self-Supervised: {Three loss terms: reconstruction, temporal consistency ($P=5$), feature consistency ($Q=5$). Weights: $\alpha=0.3$, $\beta=0.5$, reconstruction $0.2$.}
  
  \item Weakly Self-Supervised: {Two-stage training. Stage 1: same as self-supervised. Stage 2: add contrastive loss (weight $\gamma=0.8$); other weights scaled to $\alpha=0.05$, $\beta=0.1$, reconstruction $0.05$.}
\end{itemize}



\begin{adjustwidth}{-\extralength}{0cm}

\reftitle{References}

\PublishersNote{}
\end{adjustwidth}

\begin{thebibliography}{999}

\bibitem[Koch et~al.(2015)Koch, Zemel, and Salakhutdinov]{Koch2015SiameseNN}
Koch, G.; Zemel, R.; Salakhutdinov, R.
\newblock Siamese Neural Networks for One-shot Image Recognition. In Proceedings of the $32^{nd}$ International Conference on Machine
Learning, Lille, France,  6--11 July 
 2015.

\bibitem[{Chopra} et~al.(2005){Chopra}, {Hadsell}, and
  {LeCun}]{faceverification}
{Chopra}, S.; {Hadsell}, R.; {LeCun}, Y.
\newblock Learning a similarity metric discriminatively, with application to
  face verification.
\newblock In Proceedings of the 2005 IEEE Computer Society Conference on
  Computer Vision and Pattern Recognition (CVPR'05), San Diego, CA, USA, 20--25 June 2005; Volume~1, pp.
  539--546.
\newblock {\url{https://doi.org/10.1109/CVPR.2005.202}}.

\bibitem[Ruder(2017)]{MTL}
Ruder, S.
\newblock An Overview of Multi-Task Learning in Deep Neural Networks.
\newblock {\em arXiv} {\bf 2017}, { arXiv:1706.05098}.

\bibitem[Sheng and Huber(2020)]{Sheng2020Multitask}
Sheng, T.; Huber, M.
\newblock Weakly supervised multi-task representation learning for human activity analysis using wearables.
\newblock {\em Proc. ACM Interact. Mob. Wearable Ubiquitous Technol.} {\bf
  2020}, {\em 4}, 57.
\newblock {\url{https://doi.org/10.1145/3397330}}.

\bibitem[Saeed et~al.(2019)Saeed, Ozcelebi, and Lukkien]{10.1145/3328932}
Saeed, A.; Ozcelebi, T.; Lukkien, J.
\newblock Multi-task Self-Supervised Learning for Human Activity Detection.
\newblock {\em Proc. ACM Interact. Mob. Wearable Ubiquitous Technol.} {\bf
  2019}, {\em 3}, 61.
\newblock {\url{https://doi.org/10.1145/3328932}}.

\bibitem[Anguita et~al.(2013)Anguita, Ghio, Oneto, Parra, and
  L~Reyes-Ortiz]{sbhar}
Anguita, D.; Ghio, A.; Oneto, L.; Parra, X.; L~Reyes-Ortiz, J.
\newblock A Public Domain Dataset for Human Activity Recognition using
  Smartphones. In Proceedings of the ESANN 2013 Proceedings, European Symposium on Artificial Neural Networks, Computational Intelligence
and Machine Learning, Bruges, Belgium, 24--26 April 2013.

\bibitem[Reyes-Ortiz et~al.(2016)Reyes-Ortiz, Oneto, Sam\`{a}, Parra, and
  Anguita]{transitionAware}
Reyes-Ortiz, J.L.; Oneto, L.; Sam\`{a}, A.; Parra, X.; Anguita, D.
\newblock Transition-Aware Human Activity Recognition Using Smartphones.
\newblock {\em Neurocomputing} {\bf 2016}, {\em 171},~754--767.
\newblock {\url{https://doi.org/10.1016/j.neucom.2015.07.085}}.

\bibitem[Sheng and Huber(2022)]{Sheng2022consistency}
Sheng, T.; Huber, M.
\newblock Consistency Based Weakly Self-Supervised Learning for Human Activity Recognition with Wearables.
\newblock {\em In Proceedings of the AAAI-22 Workshop on Human-Centric Self-Supervised Learning, Virtual, 22 February -- 1 March 2022; HC-SSL'22.}

\bibitem[Tamura et~al.(1998)Tamura, Sekine, Ogawa, Togawa, and
  Fukui]{waveletFeature}
Tamura, T.; Sekine, M.; Ogawa, M.; Togawa, T.; Fukui, Y.
\newblock Classification of Acceleration Waveforms during Walking by Wavelet
  Transform.
\newblock {\em Methods Inf. Med.} {\bf 1998}, {\em
  36},~356--359.
\newblock {\url{https://doi.org/10.1055/s-0038-1636855}}.

\bibitem[{He} and {Jin}(2009)]{dctFeature}
{He}, Z.; {Jin}, L.
\newblock Activity recognition from acceleration data based on discrete consine
  transform and SVM.
\newblock In Proceedings of the 2009 IEEE International Conference on Systems,
  Man and Cybernetics, San Antonio, TX, USA,  11--14 October 2009; pp. 5041--5044.
\newblock {\url{https://doi.org/10.1109/ICSMC.2009.5346042}}.

\bibitem[Dua et~al.(2022)Dua, Singh, Challa, Semwal, and Sai~Kumar]{DLHAR}
Dua, N.; Singh, S.N.; Challa, S.K.; Semwal, V.B.; Sai~Kumar, M.L.S.
\newblock A Survey on Human Activity Recognition Using Deep Learning Techniques
  and Wearable Sensor Data.
\newblock In {Proceedings of the Machine Learning, Image Processing, Network
  Security and Data Sciences, Virtual,  19--20 January 2023}; Khare, N., Tomar, D.S., Ahirwal, M.K., Semwal,
  V.B., Soni, V., Eds.; Springer: Cham,  Switzerland, 2023; pp. 52--71.

\bibitem[Haresamudram et~al.(2022)Haresamudram, Essa, and
  Pl\"{o}tz]{10.1145/3550299}
Haresamudram, H.; Essa, I.; Pl\"{o}tz, T.
\newblock Assessing the State of Self-Supervised Human Activity Recognition
  Using Wearables.
\newblock {\em Proc. ACM Interact. Mob. Wearable Ubiquitous Technol.} {\bf
  2022}, {\em 6}, 116.
\newblock {\url{https://doi.org/10.1145/3550299}}.

\bibitem[Morales and Roggen(2016)]{cnnLSTMHAR}
Morales, F.J.O.; Roggen, D.
\newblock Deep Convolutional and LSTM Recurrent Neural Networks for Multimodal
  Wearable Activity Recognition.
\newblock  \emph{Sensors}  \textbf{2016}, \emph{16}, 115. 

\bibitem[{Abu Alsheikh} et~al.(2016){Abu Alsheikh}, Selim, Niyato, Doyle, Lin,
  and Tan]{DBNHAR}
{Abu Alsheikh}, M.; Selim, A.; Niyato, D.; Doyle, L.; Lin, S.; Tan, H.
\newblock Deep activity recognition models with triaxial accelerometers.
\newblock In Proceedings of the AAAI Conference on Artificial Intelligence, Phoenix, AZ, USA,  12–17 February 2016; Volume WS-16-01-WS-16-15, pp. 8--13.

\bibitem[Chen et~al.(2021)Chen, Luo, Zhao, Meng, Xie, and
  Zhu]{chen2021modeling}
Chen, R.; Luo, H.; Zhao, F.; Meng, X.; Xie, Z.; Zhu, Y.
\newblock Modeling Accurate Human Activity Recognition for Embedded Devices
  Using Multi-level Distillation.
\newblock {\em arXiv} {\bf 2021},  arXiv:2107.07331.

\bibitem[Qin et~al.(2020)Qin, Zhang, Meng, Qin, and Choo]{QIN202080}
Qin, Z.; Zhang, Y.; Meng, S.; Qin, Z.; Choo, K.K.R.
\newblock Imaging and fusing time series for wearable sensor-based human
  activity recognition.
\newblock {\em Inf. Fusion} {\bf 2020}, {\em 53},~80--87.
\newblock {\url{https://doi.org/10.1016/j.inffus.2019.06.014}}.

\bibitem[{Minh Dang} et~al.(2020){Minh Dang}, Min, Wang, {Jalil Piran}, {Hee
  Lee}, and Moon]{MINHDANG2020107561}
{Minh Dang}, L.; Min, K.; Wang, H.; {Jalil Piran}, M.; {Hee Lee}, C.; Moon, H.
\newblock Sensor-based and vision-based human activity recognition: A
  comprehensive survey.
\newblock {\em Pattern Recognit.} {\bf 2020}, {\em 108},~107561.
\newblock {\url{https://doi.org/10.1016/j.patcog.2020.107561}}.

\bibitem[Wang et~al.(2019)Wang, Chen, Hao, Peng, and Hu]{WANG20193}
Wang, J.; Chen, Y.; Hao, S.; Peng, X.; Hu, L.
\newblock Deep learning for sensor-based activity recognition: A survey. 
\newblock {\em Pattern Recognit. Lett.} {\bf 2019}, {\em 119}, 3--11. 
  {\url{https://doi.org/10.1016/j.patrec.2018.02.010}}.

\bibitem[Zhang et~al.(2022)Zhang, Li, Zhang, Shahabi, Xia, Deng, and
  Alshurafa]{dlHARsurvey}
Zhang, S.; Li, Y.; Zhang, S.; Shahabi, F.; Xia, S.; Deng, Y.; Alshurafa, N.
\newblock Deep Learning in Human Activity Recognition with Wearable Sensors: A
  Review on Advances.
\newblock {\em Sensors} {\bf 2022}, {\em 22}, 1476.
\newblock {\url{https://doi.org/10.3390/s22041476}}.

\bibitem[Nozawa and Sato(2022)]{nozawa2022evaluation}
Nozawa, K.; Sato, I.
\newblock Evaluation Methods for Representation Learning: A Survey.
\newblock In Proceedings of the Thirty-First International Joint Conference on Artificial Intelligence (IJCAI-22)
Survey Track, Vienna, Austria, 23--29 July 2022.

\bibitem[Simonyan and Zisserman(2014)]{simonyan2014very}
Simonyan, K.; Zisserman, A.
\newblock Very deep convolutional networks for large-scale image recognition.
\newblock {\em arXiv} {\bf 2014},  arXiv:1409.1556.

\bibitem[Girshick et~al.(2014)Girshick, Donahue, Darrell, and
  Malik]{girshick2014rich}
Girshick, R.; Donahue, J.; Darrell, T.; Malik, J.
\newblock Rich feature hierarchies for accurate object detection and semantic
  segmentation.
\newblock In Proceedings of the IEEE Conference on Computer
  Vision and Pattern Recognition, Columbus, OH, USA, 23--28 June 2014; pp. 580--587.

\bibitem[Sun et~al.(2017)Sun, Shrivastava, Singh, and Gupta]{sun2017revisiting}
Sun, C.; Shrivastava, A.; Singh, S.; Gupta, A.
\newblock Revisiting unreasonable effectiveness of data in deep learning era.
\newblock In Proceedings of the IEEE International
  Conference on Computer Vision, Venice, Italy, 22--29 October 2017; pp. 843--852.

\bibitem[van~der Maaten and Hinton(2008)]{tsne}
van~der Maaten, L.; Hinton, G.
\newblock Visualizing Data using {t-SNE}.
\newblock {\em J. Mach. Learn. Res.} {\bf 2008}, {\em
  9},~2579--2605.

\bibitem[McInnes et~al.(2020)McInnes, Healy, and
  Melville]{mcinnes2020umapuniformmanifoldapproximation}
McInnes, L.; Healy, J.; Melville, J.
\newblock UMAP: Uniform Manifold Approximation and Projection for Dimension
  Reduction. \emph{arXiv}  \textbf{2020}, arXiv:1802.03426.

\bibitem[Mikolov et~al.(2017)Mikolov, Grave, Bojanowski, Puhrsch, and
  Joulin]{mikolov2017advances}
Mikolov, T.; Grave, E.; Bojanowski, P.; Puhrsch, C.; Joulin, A.
\newblock Advances in pre-training distributed word representations.
\newblock {\em arXiv} {\bf 2017},  arXiv:1712.09405.

\bibitem[Kaplan et~al.(2020)Kaplan, McCandlish, Henighan, Brown, Chess, Child,
  Gray, Radford, Wu, and Amodei]{kaplan2020scaling}
Kaplan, J.; McCandlish, S.; Henighan, T.; Brown, T.B.; Chess, B.; Child, R.;
  Gray, S.; Radford, A.; Wu, J.; Amodei, D.
\newblock Scaling laws for neural language models.
\newblock {\em arXiv} {\bf 2020},  arXiv:2001.08361.

\bibitem[Carbonneau et~al.(2018)Carbonneau, Cheplygina, Granger, and
  Gagnon]{carbonneau2018multiple}
Carbonneau, M.A.; Cheplygina, V.; Granger, E.; Gagnon, G.
\newblock Multiple instance learning: A survey of problem characteristics and
  applications.
\newblock {\em Pattern Recognit.} {\bf 2018}, {\em 77},~329--353.

\bibitem[Iscen et~al.(2019)Iscen, Tolias, Avrithis, and Chum]{iscen2019label}
Iscen, A.; Tolias, G.; Avrithis, Y.; Chum, O.
\newblock Label propagation for deep semi-supervised learning.
\newblock In Proceedings of the  IEEE/CVF Conference on
  Computer Vision and Pattern Recognition, Long Beach, CA, USA,  15--20 June 2019; pp. 5070--5079.

\bibitem[Le-Khac et~al.(2020)Le-Khac, Healy, and Smeaton]{le2020contrastive}
Le-Khac, P.H.; Healy, G.; Smeaton, A.F.
\newblock Contrastive representation learning: A framework and review.
\newblock {\em IEEE Access} {\bf 2020}, {\em 8},~193907--193934.

\bibitem[Sheng and Huber(2019)]{Sheng2019SMC}
Sheng, T.; Huber, M.
\newblock Siamese Networks for Weakly Supervised Human Activity Recognition.
\newblock {In Proceedings of the 2019 IEEE International Conference on Systems, Man and Cybernetics (SMC), Bari, Italy, 6–9 October 2019.}
\newblock {\url{https://doi.org/10.1109/SMC.2019.8914045}}.

\bibitem[Zhao et~al.(2021)Zhao, Wang, Luo, Zeng, and Zha]{zhao2021self}
Zhao, Y.; Wang, G.; Luo, C.; Zeng, W.; Zha, Z.J.
\newblock Self-supervised visual representations learning by contrastive mask
  prediction.
\newblock In Proceedings of the IEEE/CVF International
  Conference on Computer Vision, Virtual, 11--17 October 2021; pp. 10160--10169.

\bibitem[Gidaris et~al.(2018)Gidaris, Singh, and
  Komodakis]{gidaris2018unsupervised}
Gidaris, S.; Singh, P.; Komodakis, N.
\newblock Unsupervised representation learning by predicting image rotations.
\newblock {\em arXiv} {\bf 2018},  arXiv:1803.07728.

\bibitem[Lee et~al.(2017)Lee, Huang, Singh, and Yang]{lee2017unsupervised}
Lee, H.Y.; Huang, J.B.; Singh, M.; Yang, M.H.
\newblock Unsupervised representation learning by sorting sequences.
\newblock In Proceedings of the  IEEE International Conference on Computer Vision, Venice, Italy, 22--29 October 2017; pp. 667--676.

\bibitem[Lee et~al.(2021)Lee, Lei, Saunshi, and ZHUO]{NEURIPS2021_02e656ad}
Lee, J.D.; Lei, Q.; Saunshi, N.; Zhuo, J.
\newblock Predicting What You Already Know Helps: Provable Self-Supervised Learning.
\newblock In \emph{Proceedings of the Advances in Neural Information Processing
  Systems}; M. Ranzato and A. Beygelzimer and Y. Dauphin and P.S. Liang and J. Wortman Vaughan, Curran Associates, Inc. 
  Virtual, 6--14 December, 2021; Volume~34, pp. 309--323.

\bibitem[Kramer(1991)]{Kramer1991NonlinearPC}
Kramer, M.A.
\newblock Nonlinear principal component analysis using autoassociative neural
  networks.
\newblock {\em Aiche J.} {\bf 1991}, {\em 37},~233--243.

\bibitem[He et~al.(2016)He, Zhang, Ren, and Sun]{resnet}
He, K.; Zhang, X.; Ren, S.; Sun, J.
\newblock Deep Residual Learning for Image Recognition.
\newblock {In Proceedings of the  2016 IEEE Conference on Computer Vision and Pattern Recognition
  (CVPR), Las Vegas, NV, USA, 27--30 June} {2016}; pp. 770--778.

\bibitem[Bromley et~al.(1993)Bromley, Guyon, LeCun, S\"{a}ckinger, and
  Shah]{signature1993}
Bromley, J.; Guyon, I.; LeCun, Y.; S\"{a}ckinger, E.; Shah, R.
\newblock Signature Verification Using a ``Siamese'' Time Delay Neural Network.
\newblock In Proceedings of the 6th International Conference
  on Neural Information Processing Systems, San Francisco, CA, USA, 29 November -- 2 December 1993; 
  NIPS'93; pp. 737--744.

\bibitem[Lea et~al.(2016)Lea, Vidal, Reiter, and Hager]{TCN1}
Lea, C.; Vidal, R.; Reiter, A.; Hager, G.D.
\newblock Temporal Convolutional Networks: {A} Unified Approach to Action
  Segmentation.
\newblock {\em arXiv} {\bf 2016}, arXiv:1608.08242.


\bibitem[Hadsell et~al.(2006)Hadsell, Chopra, and LeCun]{contrast}
Hadsell, R.; Chopra, S.; LeCun, Y.
\newblock Dimensionality Reduction by Learning an Invariant Mapping.
\newblock In Proceedings of the 2006 IEEE Computer Society Conference on
  Computer Vision and Pattern Recognition (CVPR'06),  2006, New York, NY, USA, 17--22 June 2006;
 Volume~2, pp.
  1735--1742.
\newblock {\url{https://doi.org/10.1109/CVPR.2006.100}}.

\bibitem[Vincent et~al.(2008)Vincent, Larochelle, Bengio, and
  Manzagol]{vincent2008extracting}
Vincent, P.; Larochelle, H.; Bengio, Y.; Manzagol, P.A.
\newblock Extracting and composing robust features with denoising autoencoders.
\newblock In Proceedings of the 25th International
  Conference on Machine Learning, Helsinki, Finland, 5--9 July 2008; pp. 1096--1103.

\bibitem[Kingma and Welling(2022)]{kingma2022autoencodingvariationalbayes}
Kingma, D.P.; Welling, M.
\newblock Auto-Encoding Variational Bayes. \emph{arXiv} \textbf{2022}, arXiv:1312.6114.

\bibitem[Sakurada and Yairi(2014)]{10.1145/2689746.2689747}
Sakurada, M.; Yairi, T.
\newblock Anomaly Detection Using Autoencoders with Nonlinear Dimensionality
  Reduction.
\newblock In Proceedings of the MLSDA 2014 2nd Workshop on
  Machine Learning for Sensory Data Analysis, New York, NY, USA, 2 December 2014;
  MLSDA'14; pp. 4–11.
\newblock {\url{https://doi.org/10.1145/2689746.2689747}}.

\bibitem[Kolesnikov et~al.(2019)Kolesnikov, Zhai, and
  Beyer]{DBLP:journals/corr/abs-1901-09005}
Kolesnikov, A.; Zhai, X.; Beyer, L.
\newblock Revisiting Self-Supervised Visual Representation Learning.
\newblock {\em arXiv} {\bf 2019}, {arXiv:1901.09005}.


\bibitem[Maweu et~al.(2021)Maweu, Shamsuddin, Dakshit, and
  Prabhakaran]{9421374}
Maweu, B.M.; Shamsuddin, R.; Dakshit, S.; Prabhakaran, B.
\newblock Generating Healthcare Time Series Data for Improving Diagnostic
  Accuracy of Deep Neural Networks.
\newblock {\em IEEE Trans. Instrum. Meas.} {\bf
  2021}, {\em 70}, 2508715. 
\newblock {\url{https://doi.org/10.1109/TIM.2021.3077049}}.

\bibitem[Mueller and Thyagarajan(2016)]{sentenceSiamese}
Mueller, J.; Thyagarajan, A.
\newblock Siamese Recurrent Architectures for Learning Sentence Similarity.
\newblock In Proceedings of the Thirtieth AAAI Conference on
  Artificial Intelligence, Phoenix, AZ, USA,  12--17 February 2016; AAAI Press: 	Washington, DC, U.S. 
  2016; AAAI'16; pp. 2786--2792.

\bibitem[Neculoiu et~al.(2016)Neculoiu, Versteegh, and Rotaru]{textSiamese}
Neculoiu, P.; Versteegh, M.; Rotaru, M.
\newblock Learning Text Similarity with Siamese Recurrent Networks.
\newblock In Proceedings of the Rep4NLP@ACL, Berlin, Germany,  11 August 2016.

\bibitem[Zeghidour et~al.(2016)Zeghidour, Synnaeve, Usunier, and
  Dupoux]{jointlearning}
Zeghidour, N.; Synnaeve, G.; Usunier, N.; Dupoux, E.
\newblock Joint Learning of Speaker and Phonetic Similarities with Siamese Networks.
\newblock In Proceedings of the 17th Annual Conference of the International Speech Communication Association 
San Francisco, CA, USA,  8--12 September 2016; Interspeech'16

\bibitem[Ioffe and Szegedy(2015)]{BNA}
Ioffe, S.; Szegedy, C.
\newblock Batch Normalization: Accelerating Deep Network Training by Reducing
  Internal Covariate Shift.
\newblock In Proceedings of the 32nd International
  Conference on International Conference on Machine Learning, Lille, France, 6--11 July 2015; ICML'15; Volume 37, pp. 448--456. 


\bibitem[Nair and Hinton(2010)]{RELU}
Nair, V.; Hinton, G.E.
\newblock Rectified Linear Units Improve Restricted Boltzmann Machines.
\newblock In Proceedings of the  27th International Conference on Machine Learning, Haifa, Israel, 21-24 June 2010; 
  ICML'10; pp. 807--814.

\bibitem[Lea et~al.(2016)Lea, Flynn, Vidal, Reiter, and Hager]{TCN2}
Lea, C.; Flynn, M.D.; Vidal, R.; Reiter, A.; Hager, G.D.
\newblock Temporal Convolutional Networks for Action Segmentation and
  Detection.
\newblock {\em arXiv} {\bf 2016}, {arXiv:1611.05267}.

\bibitem[Zhang et~al.(2015)Zhang, Zhao, and LeCun]{charTCN}
Zhang, X.; Zhao, J.; LeCun, Y.
\newblock Character-level Convolutional Networks for Text Classification.
\newblock In Proceedings of the 29th International
  Conference on Neural Information Processing Systems, Cambridge,
  MA, USA, 7--12 December 2015; 
 NIPS'15; Volume 1, pp. 649--657.

\bibitem[Bednarski et~al.(2022)Bednarski, Singh, Zhang, Jones, Naeim, and
  Ramezani]{bednarski2022temporal}
Bednarski, B.P.; Singh, A.D.; Zhang, W.; Jones, W.M.; Naeim, A.; Ramezani, R.
\newblock Temporal convolutional networks and data rebalancing for clinical
  length of stay and mortality prediction.
\newblock {\em Sci. Rep.} {\bf 2022}, {\em 12},~21247.

\bibitem[Chen et~al.(2020)Chen, Kornblith, Norouzi, and
  Hinton]{DBLP:journals/corr/abs-2002-05709}
Chen, T.; Kornblith, S.; Norouzi, M.; Hinton, G.E.
\newblock A Simple Framework for Contrastive Learning of Visual
  Representations.
\newblock {\em arXiv} {\bf 2020}, { arXiv:2002.05709}.

\bibitem[Xu et~al.(2023)Xu, Xie, Li, Wang, Wang, and Li]{10.1145/3593590}
Xu, L.; Xie, H.; Li, Z.; Wang, F.L.; Wang, W.; Li, Q.
\newblock Contrastive Learning Models for Sentence Representations.
\newblock {\em ACM Trans. Intell. Syst. Technol.} {\bf 2023}, {\em 14}, 67.
\newblock {\url{https://doi.org/10.1145/3593590}}.

\bibitem[Pan et~al.(2021)Pan, Wang, Wu, and Li]{pan-etal-2021-contrastive}
Pan, X.; Wang, M.; Wu, L.; Li, L.
\newblock Contrastive Learning for Many-to-many Multilingual Neural Machine
  Translation.
\newblock In Proceedings of the  59th Annual Meeting of the
  Association for Computational Linguistics and the 11th International Joint
  Conference on Natural Language Processing (Volume 1: Long Papers), Online, 1--6 August 2021; Zong, C.,
  Xia, F., Li, W., Navigli, R., Eds.; pp. 244--258.
\newblock {\url{https://doi.org/10.18653/v1/2021.acl-long.21}}.

\bibitem[Nie et~al.(2023)Nie, H.~Nguyen, Sinthong, and
  Kalagnanam]{Yuqietal-2023-PatchTST}
Nie, Y.; Nguyen, N.H.; Sinthong, P.; Kalagnanam, J.
\newblock A Time Series is Worth 64 Words: Long-term Forecasting with
  Transformers.
\newblock In Proceedings of the International Conference on Learning
  Representations, Kigali, Rwanda,  1--5 May 2023.

\bibitem[Ismail~Fawaz et~al.(2020)Ismail~Fawaz, Lucas, Forestier, Pelletier,
  Schmidt, Weber, Webb, Idoumghar, Muller, and Petitjean]{InceptionTime}
Ismail~Fawaz, H.; Lucas, B.; Forestier, G.; Pelletier, C.; Schmidt, D.F.;
  Weber, J.; Webb, G.I.; Idoumghar, L.; Muller, P.A.; Petitjean, F.
\newblock InceptionTime: Finding AlexNet for time series classification.
\newblock {\em Data Min. Knowl. Discov.} {\bf 2020}, {\em 34},~1936–1962.
\newblock {\url{https://doi.org/10.1007/s10618-020-00710-y}}.

\bibitem[Wu et~al.(2023)Wu, Hu, Liu, Zhou, Wang, and Long]{TimesNet}
Wu, H.; Hu, T.; Liu, Y.; Zhou, H.; Wang, J.; Long, M.
\newblock TimesNet: Temporal 2D-Variation Modeling for General Time Series Analysis.
\newblock In Proceedings of the International Conference on Learning Representations, Kigali Rwanda, 1--5 May 2023. ICLR'23


\bibitem[Yuan et~al.(2024)Yuan, Chan, Creagh, Tong, Acquah, Clifton, and
  Doherty]{Yuan_2024}
Yuan, H.; Chan, S.; Creagh, A.P.; Tong, C.; Acquah, A.; Clifton, D.A.; Doherty,
  A.
\newblock Self-supervised learning for human activity recognition using 700,000
  person-days of wearable data.
\newblock {\em npj Digit. Med.} {\bf 2024}, {\em 7}, 91.
\newblock {\url{https://doi.org/10.1038/s41746-024-01062-3}}.

\bibitem[Cheng et~al.(2023)Cheng, Zhang, Bu, Wu, and
  Song]{10.1016/j.knosys.2023.110789}
Cheng, D.; Zhang, L.; Bu, C.; Wu, H.; Song, A.
\newblock Learning hierarchical time series data augmentation invariances via
  contrastive supervision for human activity recognition.
\newblock {\em Know.-Based Syst.} {\bf 2023}, {\em 276}, 110789.
\newblock {\url{https://doi.org/10.1016/j.knosys.2023.110789}}.

\bibitem[Qian et~al.(2022)Qian, Tian, and Miao]{10.1145/3534678.3539134}
Qian, H.; Tian, T.; Miao, C.
\newblock What Makes Good Contrastive Learning on Small-Scale Wearable-based Tasks?
\newblock In Proceedings of the 28th ACM SIGKDD Conference on Knowledge Discovery and Data Mining, Washington DC, USA, 14--18 August 2022 
; KDD '22; pp.3761–3771.
\newblock {\url{https://doi.org/10.1145/3534678.3539134}}.

\bibitem[{Zeng} et~al.(2014){Zeng}, {Nguyen}, {Yu}, {Mengshoel}, {Zhu}, {Wu},
  and {Zhang}]{cnnHAR}
{Zeng}, M.; {Nguyen}, L.T.; {Yu}, B.; {Mengshoel}, O.J.; {Zhu}, J.; {Wu}, P.;
  {Zhang}, J.
\newblock Convolutional Neural Networks for human activity recognition using
  mobile sensors.
\newblock In Proceedings of the 6th International Conference on Mobile
  Computing, Applications and Services, Austin, TX, USA, 6--7 November 2014; pp. 197--205.

\bibitem[Hammerla et~al.(2016)Hammerla, Halloran, and
  Pl\"{o}tz]{dnnHARbenchmark}
Hammerla, N.Y.; Halloran, S.; Pl\"{o}tz, T.
\newblock Deep, Convolutional, and Recurrent Models for Human Activity
  Recognition Using Wearables.
\newblock In Proceedings of the  Twenty-Fifth International
  Joint Conference on Artificial Intelligence, New York, NY, USA, 9--15 July 2016; AAAI Press: Washington, DC, USA 
  2016; IJCAI'16; pp.
  1533--1540.

\bibitem[Sheng and Huber(2020)]{Sheng2020UnsupervisedEL}
Sheng, T.; Huber, M.
\newblock Unsupervised Embedding Learning for Human Activity Recognition Using Wearable Sensor Data.
\newblock In Proceedings of the FLAIRS Conference, Daytona Beach, FL, USA,  20--23 May 2020.

\bibitem[{Reiss} and {Stricker}(2012)]{pamap2}
{Reiss}, A.; {Stricker}, D.
\newblock Introducing a New Benchmarked Dataset for Activity Monitoring.
\newblock In Proceedings of the 2012 16th International Symposium on Wearable
  Computers, Newcastle, UK, 18--22 June 2012; pp. 108--109.
\newblock {\url{https://doi.org/10.1109/ISWC.2012.13}}.

\bibitem[Aljalbout et~al.(2018)Aljalbout, Golkov, Siddiqui, and
  Cremers]{taxonomy}
Aljalbout, E.; Golkov, V.; Siddiqui, Y.; Cremers, D.
\newblock Clustering with Deep Learning: Taxonomy and New Methods.
\newblock {\em arXiv} {\bf 2018}, {arXiv:1801.07648}.

\bibitem[Ba\~{n}os et~al.(2012)Ba\~{n}os, Damas, Pomares, Rojas, T\'{o}th, and
  Amft]{realdisp}
Ba\~{n}os, O.; Damas, M.; Pomares, H.; Rojas, I.; T\'{o}th, M.A.; Amft, O.
\newblock A Benchmark Dataset to Evaluate Sensor Displacement in Activity
  Recognition.
\newblock In Proceedings of the 2012 ACM Conference on
  Ubiquitous Computing, New York, NY, USA, 5--8 September 2012; UbiComp '12; pp. 1026--1035.
\newblock {\url{https://doi.org/10.1145/2370216.2370437}}.

\bibitem[Mahmud et~al.(2020)Mahmud, Tonmoy, Bhaumik, Rahman, Amin, Shoyaib,
  Khan, and Ali]{DBLP:journals/corr/abs-2003-09018}
Mahmud, S.; Tonmoy, M.T.H.; Bhaumik, K.K.; Rahman, A.K.M.M.; Amin, M.A.;
  Shoyaib, M.; Khan, M.A.H.; Ali, A.A.
\newblock Human Activity Recognition from Wearable Sensor Data Using
  Self-Attention.
\newblock {\em arXiv} {\bf 2020}, { arXiv:2003.09018}.

\bibitem[Liao et~al.(2022)Liao, Zhao, Liu, Ivanov, Xiong, and Yan]{Liao_2022}
Liao, T.; Zhao, J.; Liu, Y.; Ivanov, K.; Xiong, J.; Yan, Y.
\newblock Deep Transfer Learning with Graph Neural Network for Sensor-Based
  Human Activity Recognition.
\newblock In Proceedings of the 2022 IEEE International Conference on
  Bioinformatics and Biomedicine (BIBM), Las Vegas, NV, USA, 6--8 December 2022; p. 2445--2452. 
\newblock {\url{https://doi.org/10.1109/bibm55620.2022.9995660}}.

\bibitem[Tello and Degeler(2025)]{tello2025contrasting}
Tello, A.; Degeler, V.
\newblock Contrasting Global and Local Representations for Human Activity
  Recognition using Graph Neural Networks.
\newblock In Proceedings of the 40th ACM/SIGAPP Symposium on Applied Computing,
  New York, NY, USA, 31 March--4 April 2025; SAC'25.

\bibitem[Mahon and Lukasiewicz(2023)]{pmlr-v189-mahon23a}
Mahon, L.; Lukasiewicz, T.
\newblock Efficient Deep Clustering of Human Activities and How to Improve
  Evaluation.
\newblock In Proceedings of the  14th Asian Conference on
  Machine Learning, Hyderabad, India, 12--14 December 2023; Khan, E., Gonen, M., Eds.; Volume 189, pp. 722--737.  

\bibitem[Zeng et~al.(2018)Zeng, Gao, Yu, Mengshoel, Langseth, Lane, and
  Liu]{understandRNNHAR}
Zeng, M.; Gao, H.; Yu, T.; Mengshoel, O.J.; Langseth, H.; Lane, I.; Liu, X.
\newblock Understanding and Improving Recurrent Networks for Human Activity
  Recognition by Continuous Attention.
\newblock In Proceedings of the 2018 ACM International Symposium on Wearable
  Computers, New York, NY, USA, 8--12 October 2018; ISWC '18; pp. 56--63.
\newblock {\url{https://doi.org/10.1145/3267242.3267286}}.

\end{thebibliography}
\end{document}